\documentclass[lettersize,journal]{IEEEtran}

\usepackage[numbers]{natbib}
\usepackage{caption}
\usepackage{setspace}
\usepackage{amssymb}
\usepackage{multirow}
\usepackage{lscape}
\usepackage{amsthm}
\usepackage{booktabs}
\usepackage{amsmath}
\usepackage{mathrsfs}
\usepackage{longtable}
\usepackage{url}
\usepackage[table]{xcolor} 
\usepackage{tcolorbox}
\usepackage{tabularx}
\usepackage{epsfig}
\usepackage{colortbl}
\usepackage{subfigure}
\usepackage{graphicx}
\usepackage{textcomp}
\usepackage{threeparttable} 
\usepackage{array}
\usepackage{epstopdf}
\usepackage{stfloats}
\usepackage{diagbox}
\usepackage{float}
\usepackage{color}
\usepackage{algorithm}
\usepackage{algorithmicx}
\usepackage{algpseudocode}

\begin{document}

\title{Can Large Language Models Be\\ Trusted as Evolutionary Optimizers for\\ Network-Structured Combinatorial Problems?}

\author{Jie Zhao, Tao Wen, Kang Hao Cheong,~\IEEEmembership{Senior Member, IEEE}
\thanks{
This work was funded by the Agency for Science, Technology and Research (A*STAR) under the Prenatal/Early Childhood Grant (H23P1M0006).

Jie Zhao, Tao Wen and Kang Hao Cheong are affiliated with the Division of Mathematical Sciences, School of Physical and Mathematical Sciences, Nanyang Technological University, S637371, Singapore. Kang Hao Cheong is also with the School of Computer Science and Engineering, Nanyang Technological University, S639798, Singapore.

Corresponding Author: K.H. Cheong (kanghao.cheong@ntu.edu.sg).}
}

\markboth{Journal of \LaTeX\ Class Files,~Vol.~14, No.~8, August~2021}%
{Shell \MakeLowercase{\textit{\textit{et al.}}}: A Sample Article Using IEEEtran.cls for IEEE Journals}


\maketitle

\begin{abstract}
Large Language Models (LLMs) have shown strong capabilities in language understanding and reasoning across diverse domains. Recently, there has been increasing interest in utilizing LLMs not merely as assistants in optimization tasks, but as primary optimizers, particularly for network-structured combinatorial problems. However, before LLMs can be reliably deployed in this role, a fundamental question must be addressed: Can LLMs iteratively manipulate solutions that consistently adhere to problem constraints? In this work, we propose a systematic framework to evaluate the capability of LLMs to engage with problem structures. Rather than treating the model as a black-box generator, we adopt the commonly used evolutionary optimizer (EVO) and propose a comprehensive evaluation framework that rigorously assesses the output fidelity of LLM-based operators across different stages of the evolutionary process. To enhance robustness, we introduce a hybrid error-correction mechanism that mitigates uncertainty in LLMs outputs. Moreover, we explore a cost-efficient population-level optimization strategy that significantly improves efficiency compared to traditional individual-level approaches. Extensive experiments on a representative node-level combinatorial network optimization task demonstrate the effectiveness, adaptability, and inherent limitations of LLM-based EVO. Our findings present perspectives on integrating LLMs into evolutionary computation and discuss paths that may support scalable and context-aware optimization in networked systems.

\end{abstract}


\section{Introduction}

\IEEEPARstart{L}{arge} Language Models (LLMs) trained on vast datasets \cite{vaswani2017attention,brown2020language,hao2022iron} have the capability to understand and generate human-like text based on the learned patterns. LLMs excel in tasks ranging from simple text completion to complex question answering, demonstrating a nuanced understanding of language context and semantics \cite{zhou2022large,cheong2025adaptive}.  Traditional optimization is labor-intensive and demands domain expertise for precise execution \cite{wu2024evolutionary}. This challenge intensifies with constrained problems or when tailoring methods for specific needs. Such difficulty is especially pronounced for non-technical users, presenting a major barrier to accessibility and effective application. 

There is a growing interest in leveraging LLMs not merely as auxiliary tools within optimization pipelines, but as primary optimizers \cite{cai2024exploring,chao2025large}. This vision was materialized in \cite{yang2023large}, where optimization of combinatorial problems is achieved not through mathematical analysis or traditional programming but via prompt engineering only. In this work, some combinatorial problems have been used as illustrative examples, such as the traveling salesman problems (TSP), and the solution-score pair is fed into LLMs for the refined solution. Evolutionary optimization has long been recognized as a powerful approach for solving complex problems, valued for both its simplicity and effectiveness \cite{zhao2023obfuscating}. Motivated by these strengths, recent efforts have begun to explore the use of LLMs within evolutionary optimization frameworks. For example, Liu \textit{et al.} \cite{liu2024large} proposed a framework called LLM-driven evolutionary algorithm (LMEA), making LLMs function as crossover and mutation operators to solve TSP with up to 20 nodes. Brahmachary \textit{et al.} \cite{brahmachary2025large} leveraged LLMs to optimize a continuous problem with the elitism mechanism and demonstrated the potential of LLM-based optimizer. {Furthermore, Liu \textit{et al.} \cite{liu2025large} utilized LLMs to serve as black-box optimizers for decomposition-based MOEA in a zero-shot manner.} In \cite{meyerson2023language}, Meyerson \textit{et al.} explored various structures over which LLMs can perform crossover, including binary strings, mathematical expressions, natural language text, and code.

Despite these advancements, LLMs have shown variable performance across different domains; they excel in some areas, such as strategy formulation \cite{cheong2024adaptive,romera2024mathematical} but not in tasks requiring precise arithmetic and logical reasoning \cite{fatemi2023talk,wang2024can}. Therefore, to enable LLMs to reliably and robustly fulfill this role, a critical challenge must be addressed first: Can they iteratively refine and manipulate solutions while consistently preserving all domain-specific constraints throughout the optimization process? {Although encoding schemes differ considerably depending on the task, such as using permutation-based encodings for scheduling problems and binary or graph-based representations for network-related tasks, the fundamental process of manipulating these encodings remains consistent within the evolutionary algorithm framework. Core operations like crossover and mutation are applied universally, typically involving the exchange or replacement of elements, regardless of the specific encoding. Therefore, despite variations in representational strategies, evolutionary algorithms adopt a unified methodology for manipulating the solution space.}

To date, there is still a lack of comprehensive investigation to evaluate the effectiveness and reliability of LLMs as optimizers, and to identify the factors affecting their performance. Huang \textit{et al.} \cite{huang2024exploring} conducted an investigation on LLMs as normal optimizers on different problems, but it is only targeted at one-shot optimization, and some important aspects such as reliability, scalability, and computational cost remain underexplored. Due to the versatility of evolutionary optimizers on the network-related problem, it will serve the illustrative example in our work. We here aim to conduct a thorough assessment of the performance of LLMs as operators in all stages of evolutionary optimization, some of which, like LLM-based initialization and selection, still remain underexplored in the existing literature. In contrast to prior approaches that provide solution-score pairs, our method supplies only the solution to the LLM, emphasizing its ability to perform structural manipulations without explicit fitness guidance. This design reflects the modular nature of evolutionary optimization, where reproduction and evaluation are typically separated, allowing us to isolate and assess the LLM’s capacity to act as a generalizable variation operator. Moreover, it aligns with the behavior of traditional evolutionary operators, which rely solely on internal structure rather than objective values, enhancing compatibility across domains and avoiding overfitting to noisy score patterns. We aim to provide insights into the suitability of LLMs, clarifying their capabilities and limitations. 

As evolution optimization is an iterative process, an error in any step will yield a cascading effect, leading to a failure of optimization. Recognizing the inherently probabilistic nature of LLMs outputs \cite{xu2024hallucination,yao2023llm,duan2024llms} and the high requirements on the quality of the generated solutions during evolutionary optimization, we develop different stringent sets of standards for LLMs outputs in various levels to rigorously measure solutions in terms of their format, diversity, and conformity to problem constraints. We also introduce a set of corresponding error repair mechanisms with precisely tailored prompts to enhance the reliability of the LLM-based EVO.

To improve LLMs’ awareness of population-level diversity, we propose a cost-effective method that treats the entire population as the optimization unit for the LLM-based reproduction operator. We compare this population-level approach with the conventional individual-by-individual optimization method, analyzing both the quality of the generated solutions and the associated computational overhead. Some of our findings in this work are summarized as follows:

$\bullet$ \textbf{LLMs are capable of performing evolutionary operators, such as crossover, and mutation operations in terms of manipulation.} However, their effectiveness is highly sensitive to hyperparameters such as population size and solution length, which must be carefully tuned to maintain performance.

$\bullet$ \textbf{LLMs can effectively perform decision-making tasks such as selection, often outperforming traditional heuristic methods in adaptability.}  Unlike fixed-rule heuristics, LLMs can incorporate complex contextual information such as fitness values and diversity when making selection decisions.

$\bullet$ \textbf{The effectiveness and reliability of LLM-based EVO are closely tied to the capacity of the underlying foundation model.} More advanced models generally demonstrate stronger reasoning abilities, greater contextual understanding, and improved consistency across generations. 

$\bullet$ \textbf{The initialization phase is computationally intensive and might not be suitable for LLMs.} As dataset size increases, LLM's performance in this phase often degrades significantly, suggesting a need for auxiliary strategies or preprocessing.

$\bullet$ \textbf{LLMs are sensitive to the volume and complexity of input data.} When exposed to large-scale inputs, they are prone to generating infeasible or suboptimal solutions, highlighting the importance of integrated correction and repair mechanisms throughout the optimization process.

$\bullet$ \textbf{Population-level LLM-based EVO offers greater computational efficiency compared to individual-level approaches.} By operating on the entire population in a single prompt, it reduces the number of model calls and repetitive descriptions of operations. 


{The remainder of this paper is organized as follows. Section \ref{Sec.related works} reviews related work on the integration of LLMs with optimization techniques and the application of evolutionary methods to network-structured problems. Section \ref{sec.method} introduces our proposed framework for LLM-based evolutionary optimization, including design principles, operator definitions, and the repair mechanism. Section \ref{sec.experiment} presents extensive experimental evaluations, covering effectiveness, reliability, and scalability. Section \ref{Sec.dis} discusses the limitations of LLM-based EVO and outlines several promising directions for future enhancement. Section \ref{sec.conclusion} concludes the paper with a summary of findings and discussions on future directions.}

\section{Related work}\label{Sec.related works}
In this section, we will review the existing literature on the synergy of LLMs and optimization, and the application of evolutionary optimization on network-structured problems.

\subsection{Synergy of LLMs and optimization}

Yu and Liu \cite{yu2024deep} comprehensively investigated the synergy of LLMs and evolutionary optimization, and discussed different-angle applications. Wu \textit{et al.} \cite{wu2024evolutionary} classified existing works regarding the synergy of LLMs and evolutionary computation into two main branches: LLM-based black-box optimizer \cite{baumann2024evolutionary} and LLM-based algorithm automation \cite{morris2024llm,lehman2023evolution}. 

LLMs can directly serve as operators for combinatorial problems, as indicated in \cite{chao2025large}, thereby reducing the need for manual tuning and domain-specific adjustments, such as \cite{nie2023importance}. Yang \textit{et al.} proposed to use LLMs as optimizers named Optimization by PROmpting (OPRO) to solve the traveling salesman problem \cite{yang2023large}, in which the previously generated solution and its evaluation value are used as part of the prompt for the next generation. In \cite{meyerson2023language}, LLMs are employed as crossover operators to derive new solutions from parental inputs. Brownlee \textit{et al.} \cite{brownlee2023enhancing} also presented LLMs effectively functioning as mutation operators that enhance the search process. Liu \textit{et al.} \cite{liu2024large} introduced a novel framework known as LLM-driven EA (LMEA), which utilizes LLMs for both crossover and mutation operations. This approach highlights the adaptability of LLMs, where search behaviors can be easily modified by adjusting the LLMs temperatures. Furthermore, LLM-based search operators can be adapted to multi-objective scenarios by decomposing traditional optimization tasks into sub-problems \cite{liu2025large}. In \cite{wang2024large}, Wang \textit{et al.} explored the applicability of LLMs on constrained multiobjective optimization problems and achieved promising results compared to traditional methods. 

LLMs have also shown significant potential in autonomously creating and improving algorithms to effectively tackle optimization challenges \cite{huang2024autonomous}. As demonstrated in \cite{romera2024mathematical}, heuristic optimized by LLMs achieves excellent performance in different complex combinatorial problems. In \cite{liu2023algorithm}, Liu \textit{et al.} proposed a method called Algorithm Evolution using the Large Language Model (AEL), which directly treats algorithms as individuals in the evolutionary process. Then, AEL was further extended to the design of guided local search algorithms \cite{liu2024example}, showing the strength of the LLM-based method over human-designed algorithms. After that, they extended AEL to an advanced model called Evolution of Heuristics (EoH) by exploring various prompts to solve different combinatorial tasks \cite{liu2024evolution}. In addition, it was also demonstrated that LLMs can analyze swarm intelligence algorithms to obtain a hybrid algorithm that combines various strengths of existing methods \cite{pluhacek2023leveraging}. Mao \textit{et al.} explored LLM-enhanced algorithm automation for identifying critical nodes \cite{mao2024identify}. In this approach, various heuristics are initialized as populations and then evolve with the assistance of LLMs. The evolution process was also studied to enhance the adaptability and convergence by \cite{yin2024controlling}, where the mutation rate is adaptively adjusted with dynamic prompt.

LLMs can also be used as a surrogate model with the help of in-context learning to efficiently analyze the quality of the solution, as shown in \cite{hao2024large}. Furthermore, LLMs can assist in algorithm selection, as demonstrated in \cite{wu2023llm}, through analyzing the code to grasp both its structural and semantic elements, along with the contextual understanding. An emerging direction worth noting involves using evolutionary optimization to search for the optimal prompt, enabling LLMs to achieve excellent performance. Notable examples of this approach can be found in the work of \cite{li2023spell, guo2023connecting, martins2023towards}.

\subsection{Metaheuristic optimization on network-structured problems}

Network-structured combinatorial problems play a crucial role in various practical domains due to their widespread applicability \cite{zhao2025bridging,huang2024multiobjective}, such as brain analysis \cite{puxeddu2021comprehensive} and power grid resilience \cite{artime2024robustness}. The utility of evolutionary optimization in addressing discrete and non-linear problems has significantly facilitated its adoption in this field, particularly in the context of complex networks \cite{qiu2024scalable} and various combinatorial tasks \cite{wang2023multiobjective,liu2023evolutionary}. Evolutionary algorithms excel due to their inherent capability to navigate complex solution spaces effectively, making them suitable for tasks such as truck scheduling \cite{chen2024deep} and job shop scheduling \cite{yao2024bilevel,zhang2022multitask}.
In complex networks, evolutionary optimization addresses several intricate combinatorial challenges, including influence maximization \cite{liu2023identify,liu2023influence}, robustness analysis, sensor selection \cite{zhao2025enhanced,wang2021computationally,wang2023enhancing} and community deception \cite{zhao2023self,zhao2024multi}. These applications underscore the versatility and robustness of evolutionary optimization methodologies in addressing complex network problems. Notably, evolutionary algorithms have also been effectively applied in network-related tasks such as important node identification \cite{zhang2023interactive}, community discovery \cite{ma2023higher,xiao2024constrained}, network reconstruction \cite{wu2020evolutionary,ying2021multiobjective}, and network module recognition \cite{gao2023multilayer}. 


\section{LLM-based evolutionary optimizer}\label{sec.method}

For combinatorial problems in complex networks, the representation of the solution within the evolutionary optimization framework is generally defined as the index of elements:
\begin{equation}
[\text{X}_1, \text{X}_2, ..., \text{X}_n],    
\end{equation}
where $\text{X}$ depends on the specific task and $n$ is the predefined solution size \cite{chen2019ga}. In this work, we choose the influence maximization \cite{zhao2022random,tran2024meta} as an illustrative example. The problem is defined as: Given a graph $G = (V, E)$, where $V$ represents the set of nodes and $E$ represents the set of edges, the objective is to find a subset of nodes $S \subseteq V$ that maximizes the influence across the network. As for network-related problems, the solution representation can simply be 
\begin{equation}
[\text{Node}_1, \text{Node}_2, ..., \text{Node}_n].
\end{equation}
Note that our focus is not on solving any specific network-related problem but on evaluating the performance of LLMs as operators for combinatorial problems in complex networks. The reason for selecting this problem is that it is general enough to provide insights and observations for LLMs as EVO due to the representation similarity of combinatorial problems. To enhance generality, `element' will replace `node', and `dataset' will replace `graph' in the rest of the discussion.

Let $\mathbf{F}$ be combination of LLM-EVO, i.e., 
\begin{equation}
\mathbf{F} = \{\mathbf{F}_I, \mathbf{F}_S, \mathbf{F}_C, \mathbf{F}_M\},
\end{equation}
where $\mathbf{F}_I$, $\mathbf{F}_S$, $\mathbf{F}_C$ and $\mathbf{F}_M$ indicate LLM-based initialization, selection, crossover, and mutation, respectively. The prompt of LLM-based evolutionary operators and the repair strategy can be found in the supplementary material.

\begin{figure*}[!ht]
\centering
\includegraphics[width=1\textwidth]{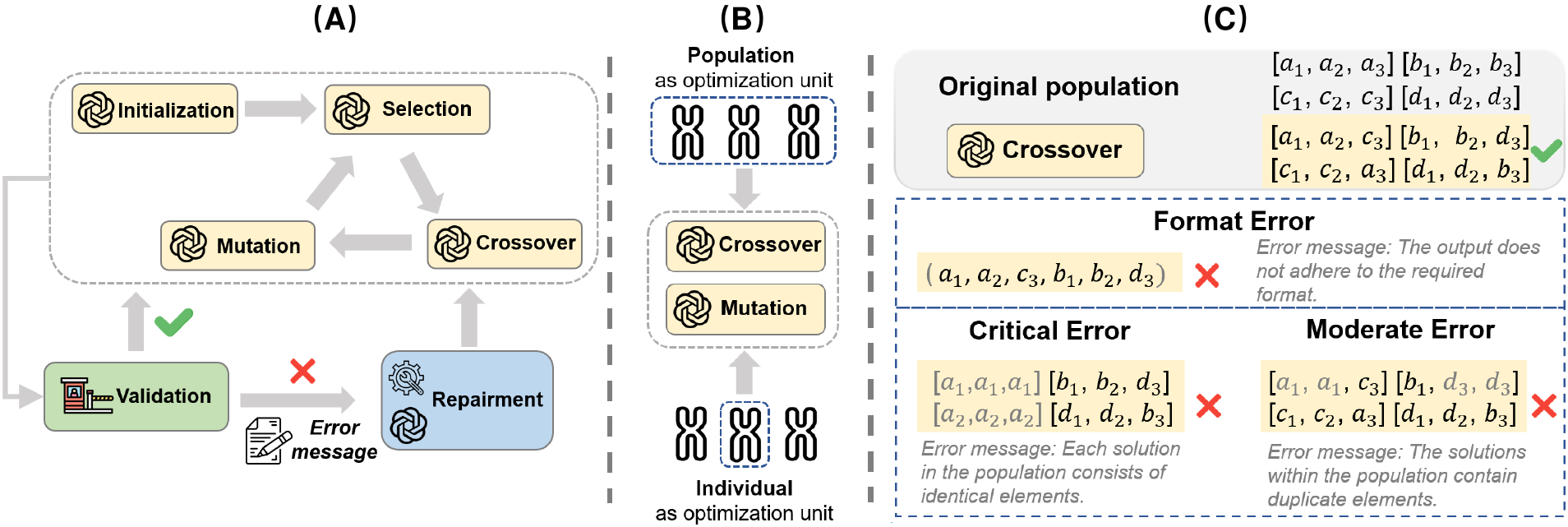}
\caption{(A) The diagram of LLM-based EVO with the proposed validation and repair mechanism. All four phases of evolutionary optimization, along with the repair process, are based on LLMs. (B) The illustration of population- and individual-level LLM-based EVO.  (C) An example of errors encountered in population-level LLM-based crossover. The error message (if any), customized corrected prompt, and previous deficient output will be provided to LLMs for repair. }
\label{diagram}
\end{figure*}

The first phase, \textbf{Initialization} is defined as $\mathcal{P}_0 = \mathbf{F}_I(G)$, where $\mathcal{P}_0$ refers to the initialized population and $G$ refers to the input data. A strategic reduction in search space can speed up convergence and improve effectiveness \cite{zhang2022search}. However, based on current research \cite{fatemi2023talk}, it is not feasible to feed datasets directly to LLMs, either in the form of adjacency matrices or text descriptions, because the reasoning ability of current LLMs on graphs remains limited.  Therefore, instead of providing LLMs with just a dataset, we opt to input element IDs along with easily accessible metrics. Following this, LLMs will be instructed to (1) Rank the elements based on a specific metric; (2) Select a percentage of top elements as candidates; and (3) Sample filtered elements to form the initialized population. To streamline the initialization process and reduce prompt complexity, we merge the ranking (1) and top-percentage selection (2) into a single step.  We separate the sampling step (3) to retain explicit control over population diversity and randomness. This decoupling enables more granular observation and analysis of each sub-process.

The \textbf{Selection} phase in evolutionary optimization is a crucial step where individuals from the current population are selected as the next generation. This is represented as $\mathcal{P}_{l+1} = \mathbf{F}_S(\mathcal{P}_l, f(\mathcal{P}_l))$, 
where $\mathcal{P}_{l+1}$ refers to the population after applying operators of $\{\mathbf{F}_C,\mathbf{F}_M\}$ on $\mathcal{P}_{l}$ and $f(\mathcal{P}_l)$ represents the fitness values of solutions in $\mathcal{P}_l$. 
Extensive studies on selection strategies, such as roulette wheel selection, aim to preferentially choose individuals with higher fitness scores. This operation allows the population to evolve towards an optimal solution over successive generations. In $\mathbf{F}_S$, we will not rely on any specified strategy but instead follow the common principles: (1) Solutions with low fitness are not allowed to be selected; and (2) Each solution can be selected multiple times but not excessively, to maintain diversity. For the selection, the input to LLMs is

\begin{equation}
\{([\text{X}_{1}^{(1)},  \ldots, \text{X}_{n}^{(1)}], f_1), \ldots, ([\text{X}_{1}^{(k)},  \ldots, \text{X}_{n}^{(k)}], f_k)\},
\end{equation}
where $k$ is the population size and $\text{X}_{i}^{(j)}$ refers to the $i$-th element in the $j$-th solution in the population. It can be simplified as
\begin{equation}
\{(S_1, f_1),...,(S_k, f_k)\},
\end{equation}
where $S_i$ is the index of solution $[\text{X}_{i}^{(1)},  \ldots, \text{X}_{n}^{(i)}]$ and $f_i$ is its fitness value.

Traditional evolutionary operators are designed to be domain-agnostic by applying consistent transformation strategies regardless of the specific objective function. By feeding only the solution into the LLM, we mimic this behavior and assess whether LLMs can fulfill the same role. The individual-level LLM-based reproduction operators of \textbf{Crossover} $\mathbf{F}_C^S$ and \textbf{Mutation} $\mathbf{F}_M^S$ are
\begin{equation}
\begin{split}
&\mathcal{P}_{l+1} = \bigcup_{\mathcal{S}_i,\mathcal{S}_j \in \mathcal{P}_l}\mathbf{F}_C^S(\mathcal{S}_i,\mathcal{S}_j),\\
&\mathcal{P}_{l+1} = \bigcup_{\mathcal{S} \in \mathcal{P}_{l}}\mathbf{F}_M^S(\mathcal{S}),
\end{split}
\label{reproduction_S}
\end{equation}
where $\mathcal{S}$ denotes the solution in population $\mathcal{P}$. The individual-level operator is the conventional method for implementing evolutionary optimization. While this individual-level method provides a precise and controlled way to reproduce the population, it may suffer from two problems: (i) Handling solutions individually or in pairs might not scale efficiently as the population size grows in the framework of LLM-based EVO; and
(ii) LLMs may lack broader contextual information about the population's overall diversity, which could impede their ability to optimize effectively. To resolve these issues, we propose a new population-level optimization, defined as follows:
\begin{equation}
\begin{split}
&\mathcal{P}_{l+1} = \mathbf{F}_C^P(\mathcal{P}_l),\\
&\mathcal{P}_{l+1} = \mathbf{F}_M^P(\mathcal{P}_l).\\
\end{split}
\end{equation}

{Here we distinguish between the two optimization paradigms: individual-level and population-level. Traditional evolutionary optimization relies on iteratively applying operators, such as crossover and mutation at the level of individual solutions. In the individual-level setting, the LLM receives one solution at a time (for mutation) or a pair of solutions (for crossover), and produces a single modified output accordingly. }

{In contrast, the population-level approach treats the entire population as a single input unit. The LLM is prompted with the full set of candidate solutions and tasked with performing crossover or mutation across all individuals in one pass. The output is an optimized population, generated in a single interaction with the model. This approach not only reduces the number of LLM calls, improving computational efficiency but also allows the LLM to consider population-wide context, such as diversity, when generating new solutions. This global awareness enables more informed and coherent optimization decisions compared to the pairwise nature of individual-level operations. The pseudocode of population- and individual-level LLM-based EVO is shown in Algorithms~\ref{alg:population-level LLM-based EVO} and \ref{alg:LLM-based EVO} respectively. }

\begin{algorithm}[!ht]
\caption{Population-level LLM-based EVO with repair mechanism} 
\hspace*{0.02in} {\bf Input:} 
Dataset $G$, fitness function $f$, maximum number of generations $N_{\text{max}}$ \\
\hspace*{0.02in} {\bf Output:} 
Optimized node set 
\begin{algorithmic}[1]
\State Initialize population $\mathcal{P} = \{\mathcal{S}_1,\mathcal{S}_2,\cdots,\mathcal{S}_n\}$ instructed by LLMs
\While{iteration count $< N_{\text{max}}$}
    \State Calculate fitness value of solutions in $\mathcal{P}$: $f(\mathcal{P})$
    \State Select solutions as $\mathcal{P}$ for reproduction instructed by LLMs
    \State Check and (repair) the output
    \State Perform crossover on $\mathcal{P}$ instructed by LLMs
    \State Check and (repair) the output
    \State Perform mutation on $\mathcal{P}$ instructed by LLMs
    \State Check and (repair) the output
\EndWhile
\State Return solution with the highest fitness value in $\mathcal{P}$.
\end{algorithmic}
\label{alg:population-level LLM-based EVO}
\end{algorithm}

\begin{algorithm}[!ht]
\caption{Individual-level LLM-based EVO with repair mechanism} 
\hspace*{0.02in} {\bf Input:} 
Dataset $G$, fitness function $f$, maximum number of generations $N_{\text{max}}$ \\
\hspace*{0.02in} {\bf Output:} 
Optimized node set
\begin{algorithmic}[1]
\State Initialize population $\mathcal{P} = \{\mathcal{S}_1,\mathcal{S}_2,\cdots,\mathcal{S}_n\}$ instructed by LLMs
\While{iteration count $< N_{\text{max}}$}
    \State Calculate fitness value of solutions in $\mathcal{P}$: $f(\mathcal{P})$
    \State Select solutions as the population for reproduction instructed by LLMs
    \State Check and (repair) the output
    \For{$S \in \mathcal{P}$}
        \State Perform crossover on $S$ instructed by LLMs
    \State Check and (repair) the output
    \State Perform mutation on $S$ instructed by LLMs
    \State Check and (repair) the output
    \EndFor
\EndWhile
\State Return solution with highest fitness value in $\mathcal{P}$.
\end{algorithmic}
\label{alg:LLM-based EVO}
\end{algorithm}

\subsection{Computational cost analysis of LLM-based EVO}
It is not applicable to analyze the complexity LLM-based optimization directly as usual as LLMs rely on online resources (e.g., token count). Here, we will discuss the difference between population- and individual-level optimization regarding computational cost.

\textbf{Crossover}:  
Let $\mathcal{V}(\cdot)$ denote the token count of its argument; thus $\mathcal{V}(P)$ and $\mathcal{V}(S)$ are the token counts of the entire population $P$ and of an individual solution $S$, respectively.  
The LLM input also contains an instruction prompt $\mathcal{T}_C^S$ (or $\mathcal{T}_C^P$).  
These two prompts are almost equal in length, i.e.,
\[
\mathcal{V}(\mathcal{T}_C^P)=\mathcal{V}(\mathcal{T}_C^S)+\delta_C,
\qquad
\delta_C\ll\mathcal{V}(\mathcal{T}_C^S),
\]
where $\delta_C$ is the extra administrative phrase for population-level manner, such as
\textit{“Please randomly select pairs of solutions and continue applying the crossover operation until the number of newly created solutions matches the predefined population size.”}

Regarding the overall cost, we can have

\emph{Population-level cost}:  
\[
\mathcal{V}(\mathcal{T}_C^P)+\mathcal{V}(P).
\]

\emph{Individual-level cost}:  
\[
\bigl(\mathcal{V}(\mathcal{T}_C^S)+2\,\mathcal{V}(S)\bigr)\,\frac{\mathcal{N}_p}{2},
\]
where $\mathcal{N}_p$ is the population size.

Because $\mathcal{V}(P)=\mathcal{N}_p\,\mathcal{V}(S)$, the cost difference is  
\begin{align*}
\Delta_C 
&= \left( \mathcal{V}(\mathcal{T}_C^S) + 2\,\mathcal{V}(S) \right) \cdot \frac{\mathcal{N}_p}{2}
   - \left( \mathcal{V}(\mathcal{T}_C^S) + \delta_C + \mathcal{N}_p\,\mathcal{V}(S) \right) \\
&= \frac{\mathcal{N}_p}{2}\,\mathcal{V}(\mathcal{T}_C^S) + \mathcal{N}_p\,\mathcal{V}(S)
   - \mathcal{V}(\mathcal{T}_C^S) - \delta_C - \mathcal{N}_p\,\mathcal{V}(S) \\
&= \left( \frac{\mathcal{N}_p}{2} - 1 \right) \mathcal{V}(\mathcal{T}_C^S) - \delta_C \\
&\approx \frac{\mathcal{N}_p - 2}{2} \, \mathcal{V}(\mathcal{T}_C^S).
\end{align*}

\textbf{Mutation}:  
Define $\mathcal{V}(\mathcal{T}_M^P)=\mathcal{V}(\mathcal{T}_M^S)+\delta_M$ analogously.

\emph{Population-level cost}:  
\[
\mathcal{V}(\mathcal{T}_M^P)+\mathcal{V}(P).
\]

\emph{Individual-level cost}:  
\[
\bigl(\mathcal{V}(\mathcal{T}_M^S)+\mathcal{V}(S)\bigr)\,\mathcal{N}_p.
\]

Subtracting, and again using $\mathcal{V}(P)=\mathcal{N}_p\,\mathcal{V}(S)$, gives  
\[
\Delta_M
   =\mathcal{N}_p\,\mathcal{V}(\mathcal{T}_M^S)-\delta_M
   \approx(\mathcal{N}_p-1)\,\mathcal{V}(\mathcal{T}_M^S).
\]

\textbf{Overall saving}:  
\[
\Delta_C+\Delta_M
\;\approx\;
\frac{\mathcal{N}_p-2}{2}\,\mathcal{V}(\mathcal{T}_C^S)
+(\mathcal{N}_p-1)\,\mathcal{V}(\mathcal{T}_M^S).
\]

Therefore, the proposed population-level method saves roughly  
\(
\frac{\mathcal{N}_p-2}{2}\,\mathcal{V}(\mathcal{T}_C^S)+
(\mathcal{N}_p-1)\,\mathcal{V}(\mathcal{T}_M^S)
\)  
tokens in each evolutionary round compared with the individual-level method, and this advantage grows linearly with the population size and the number of rounds.

\begin{table*}[!ht]
\centering
\caption{
Categorization and description of format, critical, and moderate errors (E1–E15) used to evaluate LLM-generated outputs during different phases of the evolutionary optimization process.}
\begin{tabular}{lcl}
\hline
\textbf{Error Type} & \textbf{Error Index} & \textbf{Error Description} \\
\hline
\multirow{2}{*}{\textbf{Format Error}} 
& \textbf{E1} & The output is not in the required format. \\
& \textbf{E2} & The output contains non-integer elements. \\
\hline
\multirow{7}{*}{\textbf{Critical Error}} 
& \textbf{E3} & The selected candidates significantly deviate from the ground truth. \\
& \textbf{E4} & The size of candidates falls significantly short of meeting the requirements. \\
& \textbf{E5} & The size of the population falls significantly short of meeting the requirements. \\
& \textbf{E6} & The selected population contains one solution too many times. \\
& \textbf{E7} & Any solution appears in the population where all elements are the same. \\
& \textbf{E8} & The number of different elements in the solution changes significantly. \\
& \textbf{E9} & The number of different solutions in the population changes significantly. \\
\hline
\multirow{6}{*}{\textbf{Moderate Error}} 
& \textbf{E10} & The size of some solutions fails to meet the requirement. \\
& \textbf{E11} & The solution (population) after the operation remains the same as before. \\
& \textbf{E12} & The size of the population fails to meet the requirement. \\
& \textbf{E13} & The new solution contains duplicated elements. \\
& \textbf{E14} & The new solution contains invalid elements not found in candidate nodes. \\
& \textbf{E15} & The selected population contains those with very low fitness. \\
\hline
\end{tabular}
\label{error_message}
\end{table*}

\subsection{Error repair}

\begin{table*}[!ht]
\centering
\caption{Required error checks for LLM output across evolutionary optimization phases
Checklist of applicable format (E1–E2), critical (E3–E9), and moderate (E10–E15) error validations per optimization phase.}
\begin{tabular}{cccccccccccccccc}  
\toprule  
\multirow{2}{*}{\textbf{Phase}}&  
\multicolumn{2}{c}{\textbf{Format Error}}&\multicolumn{7}{c}{\textbf{Critical Error}}&\multicolumn{5}{c}{\textbf{Moderate Error}}\cr  
\cmidrule(lr){2-3} \cmidrule(lr){4-10} \cmidrule(lr){11-16} 
& \textbf{E1} & \textbf{E2} & \textbf{E3} & \textbf{E4} & \textbf{E5} & \textbf{E6} & \textbf{E7} & \textbf{E8} & \textbf{E9} & \textbf{E10} & \textbf{E11} & \textbf{E12} & \textbf{E13} & \textbf{E14} & \textbf{E15}\cr
\midrule  
\textbf{Candidate Selection} & \checkmark& \checkmark& \checkmark& \checkmark& -& -& -& -& -& -& -& -& -& -& -\\
\textbf{Initialization} & \checkmark& \checkmark& -& -& -& -& \checkmark& -& -& \checkmark& -& \checkmark& \checkmark& \checkmark& -\\
\textbf{Selection} & \checkmark& \checkmark& -& -& -& \checkmark& -& -& -& -& -& \checkmark& -&  - &\checkmark\\
\textbf{Crossover (P)}& \checkmark& \checkmark& -& -& \checkmark& -& \checkmark& \checkmark& \checkmark& \checkmark& \checkmark& \checkmark& \checkmark& -& -\\
\textbf{Crossover (S)}& \checkmark& \checkmark& -& -& -& -& \checkmark& \checkmark& -& \checkmark& \checkmark& -& \checkmark& -& -\\
\textbf{Mutation (P)}& \checkmark& \checkmark& -& -& \checkmark& -& \checkmark& \checkmark& \checkmark& \checkmark& \checkmark& \checkmark& \checkmark& \checkmark& -\\
\textbf{Mutation (S)}& \checkmark& \checkmark& -& -& -& -& \checkmark& \checkmark& -& \checkmark& \checkmark& -& \checkmark& \checkmark& -\\ 
\bottomrule  
\end{tabular}  
\label{phase_check}
\end{table*}

The probabilistic nature of LLMs can lead to occasional undesirable results. As evolutionary optimization is an iterative process, an error early in the process can `blow up' in subsequent cycles, creating a cascading effect to ruin the optimization process. To investigate the reliability of LLM-generated outputs across different phases, we establish rigorous validation standards and systematically investigate potential errors encountered during simulation. These errors are categorized into format errors and quality errors, with the latter further divided into critical and moderate types.

The summary of the different types of errors is listed in Table \ref{error_message} and details are introduced as follows. 

\textbf{Format Error}:
The format error refers to those that can directly interrupt the running of a program. Since format errors are rare and difficult to fix, it is usually more efficient to request a new generation.

\begin{itemize}
\item \textbf{E1: The output is not in the required format.} 
\begin{itemize}
\item \textbf{Example:} When we ask LLMs to generate a list in Python, we expect only that list without any explanation and text for further operations. However, LLMs sometimes add extra text, like "\textit{You are doing a crossover, and the result is ...}".
\item \textbf{Impact:} This error can disrupt the optimization process because algorithms depend on a specific data structure.
\end{itemize}

\item \textbf{E2: The output contains non-integer elements.} 
\begin{itemize}
\item \textbf{Example:} The combinatorial problems require discrete integer values while LLMs sometimes output a solution containing non-integer numbers such as [3, 5.9, 2.1, 8].

\item \textbf{Impact:} Non-integer outputs can lead to invalid candidate solutions that cannot be evaluated, thus this error will also interrupt the process as E1.

\end{itemize}

\end{itemize}

\textbf{Critical Error:}
Most critical errors undermine diversity, which is essential for successful optimization. Note that once diversity is severely damaged, then it is very difficult to repair as only mutation contributes to exploration in the entire evolutionary optimization. Therefore, this kind of error is not allowed in optimization. In a manner consistent with addressing format errors, a new output will be requested instead of attempting to correct an unacceptable one.

\begin{itemize}
\item \textbf{E3: The selected candidates significantly deviate from the ground truth.} 
\begin{itemize}
\item \textbf{Example:} We input all possible elements to LLMs to filter, each with a specific metric, and seek the top 50\% (for example). However, LLMs outputs may only minimally overlap with the ground truth, indicating that most of their suggestions are not suitable candidates. 

\item \textbf{Impact:} This case will reduce the quality of the initialized population, further impacting subsequent optimization.

\end{itemize}

\item \textbf{E4: The size of candidates falls significantly short of meeting the requirements.} 
\begin{itemize}
\item \textbf{Example:} We input all possible elements, each with a specific metric, and seek the top 50\%. However, LLMs may output only 10\% of the elements.

\item \textbf{Impact:} Such a number is insufficient to guarantee the diversity of the population as the solutions in the initialized population will be very similar.

\end{itemize}

\item \textbf{E5: The size of the population falls significantly short of meeting the requirements.} 
\begin{itemize}
\item \textbf{Example:} We input the entire population consisting of 30 solutions into LLMs but only receive 10 solutions as output.

\item \textbf{Impact:} The drastic reduction in population size will severely affect the diversity of the population. The population size may be restored to the predefined number but the diversity remains poor as there will be a lot of repetitive solutions.

\end{itemize}

\item \textbf{E6: The selected population contains one solution too many times.} 
\begin{itemize}
\item \textbf{Example:} We input a set of solution IDs and wish to filter those low-fitness solutions. However, LLMs may occasionally produce repetitive IDs for numerous identical high-fitness solutions.

\item \textbf{Impact:} Although the low-fitness solutions are filtered, over-selecting a single solution damages the diversity of the population. 
\end{itemize}

\item \textbf{E7: Any solution appears in the population where all elements are the same.} 
\begin{itemize}
\item \textbf{Example:} LLMs sometimes return  solutions like $[A_1, A_1, A_1, A_1]$ but the constraint is that no repetitive element is allowed within the same solution.

\item \textbf{Impact:} When this kind of error occurs, we discovered that the number of erroneous solutions in the population is not a few but a lot. After several rounds of crossover, all solutions of the entire population will be polluted, resulting in poor diversity.

\end{itemize}

\item \textbf{E8: The number of different elements in the population changes significantly.} 
\begin{itemize}
\item \textbf{Example:} When we input a population containing 50 different elements, LLMs sometimes will output a population containing only 20 elements.

\item \textbf{Impact:} The drastic reduction in the number of different elements will severely affect the diversity of the population.

\end{itemize}

\item \textbf{E9: The number of different solutions in the population changes significantly.} 
\begin{itemize}
\item \textbf{Example:} This error is similar to E6 that occurs during selection, whereas this error occurs during reproduction. When we input a population consisting of 30 various solutions to LLMs, the output LLMs may also contain 30 solutions but most of which may be identical.

\item \textbf{Impact:} The drastic reduction in the number of different solutions will severely affect the diversity of the population.

\end{itemize}

\end{itemize}

\textbf{Moderate Error:}
The moderate errors have a high chance of being repaired during the optimization thus they will not yield severe impact.

\begin{itemize}
\item \textbf{E10: The size of some solutions fails to meet the requirement.} 
\begin{itemize}
\item \textbf{Example:} When we input a solution consisting of 10 elements to LLMs for reproduction, the output may have 9 or 11 elements.

\end{itemize}

\item \textbf{E11: The solution (population) after the operation remains the same as before.} 
\begin{itemize}
\item \textbf{Example:} When we input a solution (population) to LLMs for reproduction, the output remains unchanged. For example, the solution $[A_1, A_2, A_3, A_4]$ is still $[A_1, A_2, A_3, A_4]$ after mutation.

\end{itemize}

\item \textbf{E12: The size of the population fails to meet the requirement.} 
\begin{itemize}
\item \textbf{Example:} When we input a population consisting of 30 solutions for reproduction, LLMs may output a population containing 28, 29, 31, or 32 solutions.
\end{itemize}

\item \textbf{E13: The new solution contains duplicated elements.} 
\begin{itemize}
\item \textbf{Example:} The issue resembles E7, but is less severe and involves only 2 or 3 identical elements in the output solution.
\end{itemize}

\item \textbf{E14: The new solution contains invalid elements not found in candidate nodes.} 
\begin{itemize}
\item \textbf{Example:} Suppose we have 100 elements and filter the 50 as candidates, the LLMs sometimes output a solution containing the element in the other 50. We found that LLMs did not produce elements outside these 100, thus we categorize this error as moderate.
\end{itemize}

\item \textbf{E15: The selected population contains those with very low fitness.} 
\begin{itemize}
\item \textbf{Example:} This error is similar to E11, but it pertains specifically to selection. When we provide a list of solutions to an LLM for selection, it tends to reproduce the same input solutions, resulting in low-fitness options being retained.
\end{itemize}

\end{itemize}

An example of possible errors encountered in the population-level crossover is given in Figure \ref{diagram}(C). The outputs generated in different phases will undergo different validations due to the varying requirements for each phase. In addition, the output requirements are different when the input is an individual solution (denoted as $S$) and the entire population (denoted as $P$). The detail is shown in Table \ref{phase_check}.

During optimization, particularly in environments with numerous constraints, managing invalid solutions is critical in ensuring successful optimization. To this end, we introduce a robust repair mechanism that maintains solution feasibility and ensures that each iteration contributes positively toward finding an optimal solution. Let $\mathcal{O}$ be the output from $\{\mathbf{F}_I, \mathbf{F}_S, \mathbf{F}_C, \mathbf{F}_M\}$, the refined output is defined as
\begin{equation}
\mathcal{O} \gets \mathbf{F}_R(\mathcal{O},E_x, R_x),
\end{equation}
where $\mathbf{F}_R$ refers to the LLM-based repair operator. $E_x$ and $R_x$ denote the error message and targeted repair prompt.

During the optimization, each output undergoes the three aforementioned examinations. If any error is detected, the repair mechanism is triggered. We will check and repair each type of error individually. Format errors and critical errors are particularly detrimental. The former interrupts the entire optimization process, and the latter can heavily corrupt the population, diverting the optimization from its optimal path. We will avoid repairing these two types of errors due to the complexity involved. Instead, a new generation is directly requested to obtain a valid solution. Due to their destructive impact, solutions failing these checks cannot be used and the previous phase's solution will be used for the next phase.

In contrast, solutions with moderate errors that are not successfully repaired are allowed to enter into the next phase. This approach is adopted because moderate errors do not yield cascading effects on the entire optimization, and there will still be an opportunity to repair them in the next phases. For the process of validation and repair, please refer to Algorithm \ref{alg:check_repair_2}. In each stage, the output of LLMs will undergo the corresponding check and repair (if any). For the checklist of each stage, please refer to Table 2 of the main text.

\begin{algorithm}[!ht]
\caption{Repair mechanism for iterative optimization}
\label{alg:check_repair_2}
\begin{algorithmic}[1]
\Require Output $\mathcal{O}_x$ obtained by $\mathbf{F}_x \in \{\mathbf{F}_I, \mathbf{F}_S, \mathbf{F}_C, \mathbf{F}_M\}$
\Ensure Refined output
\State Retrieve the checklist $\mathbf{L}_F$ in the phase regarding $\mathbf{F}_x$
\For{each $E_x$ in $\mathbf{L}_F$}
\State Check the input regarding $E_x$
\If{$E_x$ passed}
\State $\mathcal{O}_x \gets \mathcal{O}_x$ \Comment{Output $\mathcal{O}_x$ remains unchanged}
\Else
\State Retrieve the corresponding repair prompt $R_x$
\State $\mathcal{O}_x \gets \mathcal{F}_R(\mathcal{O}_x,E_x,R_x)$ \Comment{pass the repaired result to the next check}
\EndIf
\EndFor\\
Return the final output $\mathcal{O}_x$
\end{algorithmic}
\end{algorithm}

\subsection{Solution evaluation}

After the output undergoes the check and, if necessary, repair process, we categorize it into the following cases:
\begin{itemize}
    \item \textbf{Approved ($\mathsf{Q}_\mathsf{app}$):} The solution meets all requirements and standards perfectly, with no errors or deficiencies. It is ready for implementation without any modifications.
    \item \textbf{Repaired ($\mathsf{Q}_\mathsf{rep}$):} The solution had minor issues that did not meet the necessary standards, but these have been addressed, and it now meets the required criteria.
    \item \textbf{Acceptable ($\mathsf{Q}_\mathsf{acc}$)}: The solution contains flaws that fail to be completely corrected, but it still functions adequately and meets the minimum necessary criteria for use, albeit not optimally.
    \item \textbf{Rejected ($\mathsf{Q}_\mathsf{rej}$):} The solution has format or critical error even though after an attempt of repair, rendering it completely unusable.
\end{itemize}
For the format and critical errors, the acceptable case $\mathsf{Q}_\mathsf{acc}$ is not applicable due to their destructive impact on the population. If a solution encounters these errors and cannot be repaired, it must be rejected. On the contrary, the output with moderate error from the last check will pass through to the next phase if repair fails.

\section{Experimental studies}\label{sec.experiment}
In this section, we will examine the ability of LLMs to manipulate the solution of the network-structured problems.

\subsection{Experimental settings and dataset}
The fidelity and reliability of LLM-based EVO of different stages are validated on various datasets in different settings. Each simulation uses a fixed population size of 30 and runs the evolutionary process for 30 generations. During initialization, 50\% of the nodes are selected as candidates based on their degree centrality, which serves as the metric for candidate filtering. {The results are averaged from 10 independent simulations.}  To ensure a fair comparison between the population-level and individual-level LLM-based optimization, we set both the crossover and mutation rates to 1.0 in the individual-level setting, ensuring that every solution is subject to reproduction in each generation. The performance was evaluated across three language models, i.e., GPT-3.5, GPT-4.0, and GPT-4o (used specifically for testing the repair mechanism) with the temperature parameter set to 0.8 to balance output diversity and generation stability. 
Table~\ref{tab.topological} provides structural details about the tested networks, and the data is available online\footnote{http://www-personal.umich.edu/mejn/netdata.}.

\begin{table}[!ht]
\centering
\caption{Topological information of networks, including $|V|$ and $|E|$ for the number of nodes and edges, respectively, $\langle K\rangle$ for the average degree, and CC and ASD for the clustering coefficient and average shortest distance.} 
\begin{tabular}{lcccccc}
\hline \textbf{Network} & $|V|$ & $|E|$ & $\langle K\rangle$ &  CC & ASD  \\
\hline
\textbf{Dolphins} & 62 & 159 & 5.13  & 0.308 & 3.454 \\
\textbf{Netscience} & 379 & 914 & 4.82  & 0.741 & 6.061 \\
\textbf{Erods} & 433 & 1,314 & 6.06  & 0.347 & 4.021\\
\textbf{Email}& 1,005 & 25,571 & 50.89  & 0.267 & 2.587 \\
\textbf{Astro} & 14,845 & 119,652 & 16.12  & 0.425 & 4.798 \\
\hline
\end{tabular}
\label{tab.topological}
\end{table}

\subsection{Fitness function} 

In this study, we use the problem influence maximization that is extensively explored by the evolutionary computation community as an illustration \cite{wen2024eriue,wang2019surrogate} due to its generalizability. Given a graph $G = (V, E)$, where $V$ represents the set of nodes and $E$ represents the set of edges, the objective is to find a subset of nodes $S \subseteq V$ that maximizes the influence across the network. Let $\{C_1, C_2, \ldots, C_k\}$ be the communities partition. The overall fitness is empirically computed as a weighted sum of the influence within each community:
\begin{equation}
f(S) = \sum_{i=1}^k \frac{|C_i|}{|V|} \cdot |I(S) \cap C_i|,
\end{equation}
where each community's weight is proportional to its size relative to the total number of nodes. $I(S)$ refers to the set of influenced nodes within 2 hops from the seed nodes.

\subsection{Validation of LLM-based EVO}

\subsubsection{Initialization}\label{app.Initialization}
Table \ref{candidate_error} shows the results of the pass ratio of format and critical checks in the candidate selection, consisting of ranking and filtering. As observed, LLMs that perform well on small datasets often fail to generate valid outputs when applied to larger datasets. For example, on the Dolphins dataset, LLMs achieve 100.0\% $\mathsf{Q}_\mathsf{app}$ on both checks. In contrast, when applied to larger networks, we observe a clear decline in pass ratios particularly, the critical check pass drops to about 40\%. Thus, it can be seen that \textbf{LLMs performance in calculation tasks, like candidate selection, will decline as the amount of input data increases.} 

\textbf{While LLMs possess the potential for global awareness, enabling them to generate a more diverse and well-distributed initial population, practical limitations arise when applied to large-scale networks.} To make informed decisions, LLMs require structured input that includes node IDs along with relevant metrics, such as degree, betweenness centrality, or clustering coefficients. However, encoding this information for all nodes in a large graph can quickly exceed the model’s token limit, making full-graph initialization infeasible. This token overhead significantly restricts scalability, necessitating pre-filtering strategies or metric-based candidate selection before LLMs engagement.

\begin{figure}[!ht]
\centering
\includegraphics[width=0.49\textwidth]{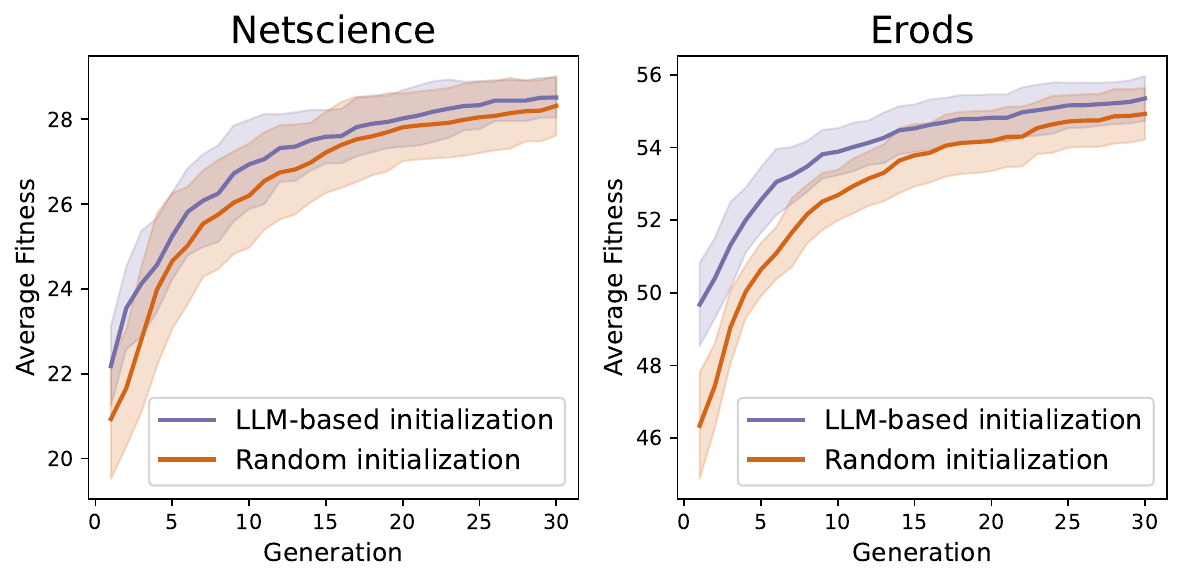}
\caption{Average fitness over generations using LLM-based initialization versus random initialization on the Netscience and Erods datasets. The backbone LLM is GPT-4.0.}
\label{fig.Initialization_fitness}
\end{figure}

\begin{table}[!ht]
\centering
\caption{Validation results of LLM-Based candidate selection. Pass rates of format and critical checks for LLM-generated candidates across three networks.  The backbone LLM is GPT-4.0.}
\begin{tabular}{ccccc}
\toprule  
\textbf{Dataset}& \textbf{Dolphins}& \textbf{Netscience} & \textbf{Erods} \\
\midrule  
\textbf{Format check pass}& 100.0\%& 95.0\%& 94.0\%\\
\textbf{Critical check pass}& 100.0\%& 41.0\%& 42.0\%\\
\bottomrule 
\end{tabular}
\label{candidate_error}
\end{table}
Therefore, we restrict our evaluation of LLM-based initialization to small networks as a preliminary study, focusing only on cases where the population is successfully initialized. As shown in Figure~\ref{fig.Initialization_fitness}, the LLM-based approach yields higher initial fitness and maintains superior performance throughout the optimization compared to random initialization. These results demonstrate that LLMs can effectively manage population initialization for networks comprising hundreds of nodes.

\subsubsection{Selection} 
The following selection strategies are used as baselines for comparison with the LLM-based approach. \textbf{Random selection} chooses individuals randomly as parents for the next generation, without considering their fitness. \textbf{No selection} means every individual in the population survives to the next generation, regardless of fitness. \textbf{Roulette selection} assigns selection probabilities based on fitness, with fitter individuals having a higher chance of being selected. \textbf{Tournament selection} involves selecting a subset of individuals randomly and choosing the best from this subset as a parent. 

\begin{figure*}[!ht]
\centering
\includegraphics[width=0.7\textwidth]{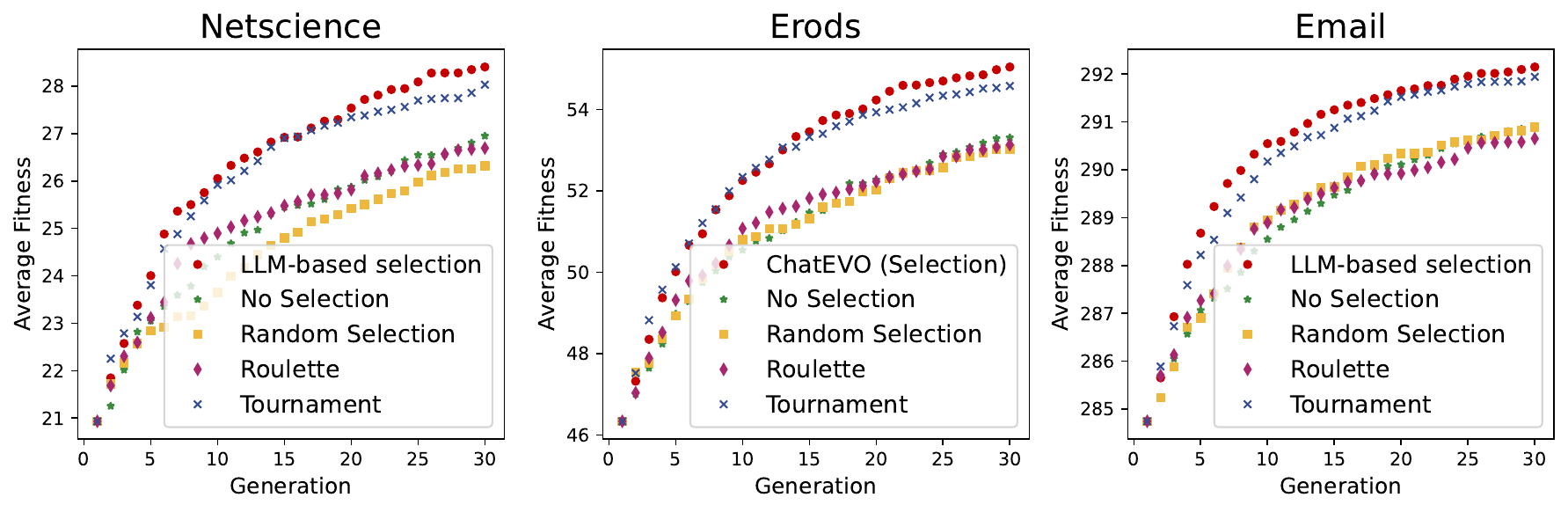}
\caption{{Average fitness over generations for LLM-based, no selection, random selection, roulette, and tournament strategies on Netscience, Erods, and Email datasets. The backbone LLM is GPT-4.0.}}
\label{fig.selection_fitness}
\end{figure*}

Figure~\ref{fig.selection_fitness} compares the selection performance of LLM-based selection with other selection strategies. The LLM-based selection demonstrates the highest fitness across the entire optimization generation, with only the tournament strategy achieving comparable performance. It can be deduced that \textbf{LLMs are effective in decision-making tasks such as in the selection phase.} {This strong performance can be attributed to the LLMs’ ability to perform adaptive selection that balances exploitation and exploration. In contrast to static rule-based methods that rely on predefined selection probabilities, LLMs offer flexibility by dynamically adjusting their selection criteria in response to the optimization process. This adaptability enables LLMs to sustain steady optimization progress while reducing the risk of premature convergence.}

\begin{figure*}[!ht]
\centering
\includegraphics[width=0.7\textwidth]{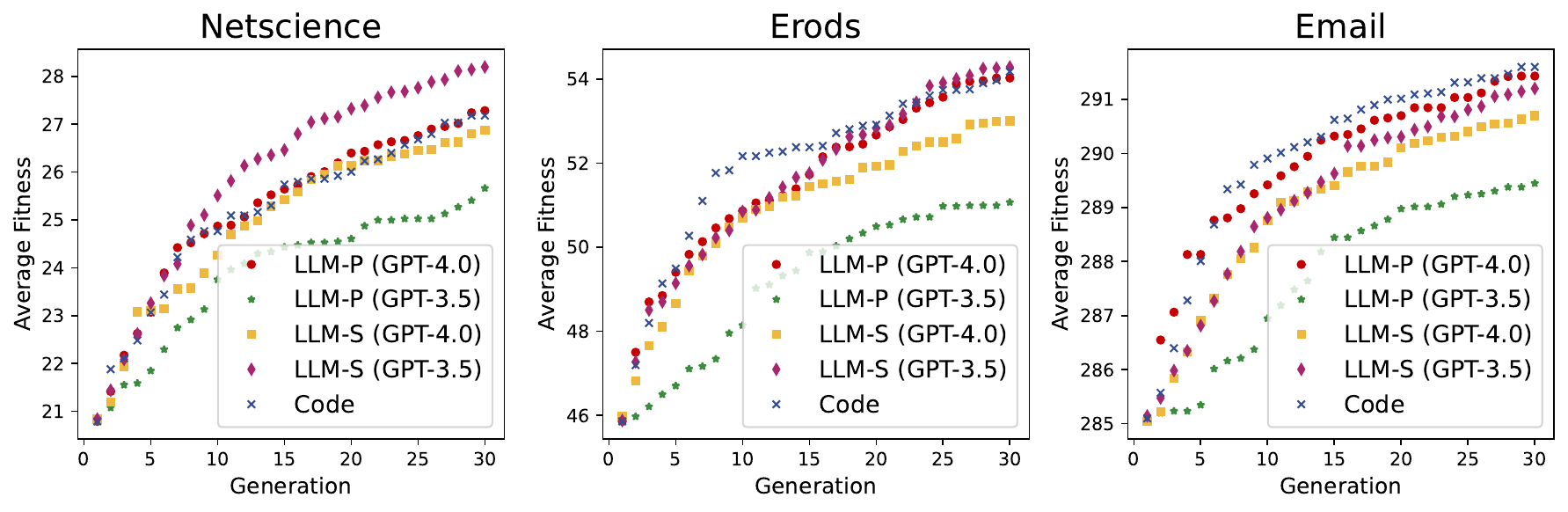}
\caption{{
Average fitness over generations for population-level (LLM-P) and individual-level (LLM-S) LLM-based optimizers using GPT-4.0 and GPT-3.5, compared to code-based optimization on Netscience, Erods, and Email datasets.}}
\label{fig.compare_fitness}
\end{figure*}

To ensure a fair comparison, we enhanced the baseline soft-based evolutionary optimizer (probability-based EVO) used in Figure~\ref{fig.compare_fitness} by enforcing strict constraint satisfaction. As shown, \textbf{the LLM-based EVO achieves performance comparable to that of traditional software-based optimizers}, as evidenced by their similar fitness values throughout the optimization process. This suggests that, in this specific setting, LLMs are capable of accurately manipulating candidate solutions during reproduction, including tasks such as crossover and mutation. Furthermore, the results highlight that model choice plays a critical role: the population-level LLM-based EVO using GPT-3.5 significantly underperforms, indicating the importance of using sufficiently capable models to ensure solution quality.

In addition, our results show that population-level optimization outperforms individual-level optimization when using GPT 4.0. This result suggests that \textbf{providing the entire population as input enables the model to leverage global context, resulting in more diverse and superior offspring.} These findings provide strong motivation for adopting LLMs as evolutionary operators, offering a compelling alternative to traditional probability-based methods that typically operate on individuals in isolation.

\begin{figure}[!ht]
\centering
\includegraphics[width=0.49\textwidth]{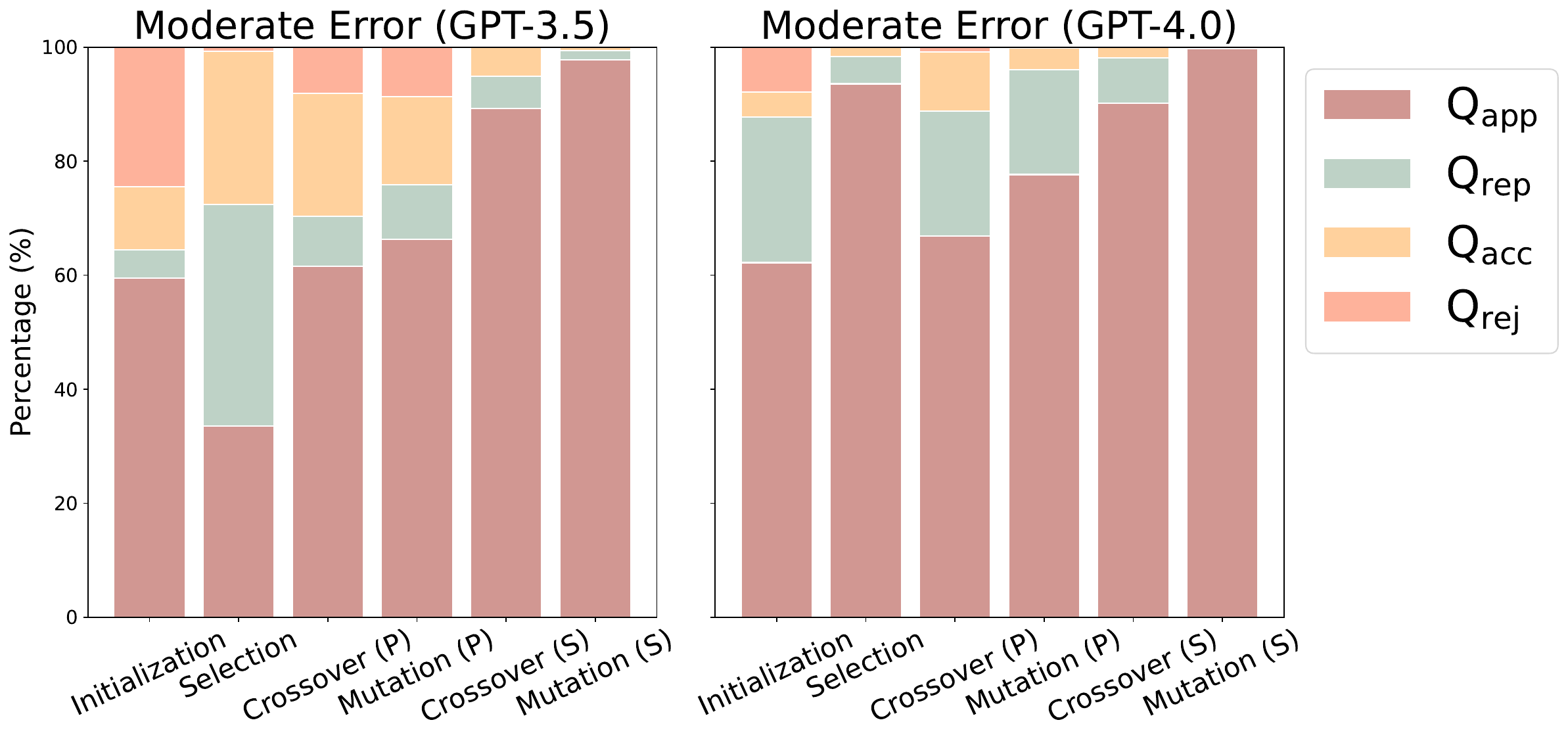}
\caption{Distribution of moderate error outcomes across evolutionary phases for GPT-3.5 and GPT-4.0 generated outputs.}
\label{fig.netscience_moderate_error}
\end{figure}

\subsection{Reliability analysis}

\begin{table*}[!ht]
\setlength{\tabcolsep}{1.3mm} 
\centering
\caption{The validations for the output generated in different phases. The tested network is Netscience. (P) refers to population-level reproduction and (S) refers to individual-level reproduction. Initialization refers to the sampling procedure.}
\resizebox{\textwidth}{!}{
\begin{tabular}{ccccccccccccc}  
\toprule  
\multirow{2}{*}{\textbf{Netscience}}&  
\multicolumn{3}{c}{\textbf{Format Error} (GPT-4.0)}&\multicolumn{3}{c}{\textbf{Critical Error} (GPT-4.0)}&\multicolumn{3}{c}{\textbf{Format Error} (GPT-3.5)}&\multicolumn{3}{c}{\textbf{Critical Error} (GPT-3.5)}\cr  
\cmidrule(lr){2-4} \cmidrule(lr){5-7} \cmidrule(lr){8-10}\cmidrule(lr){11-13}  
& $\boldsymbol{\mathsf{Q}_\mathsf{app}}$ & $\boldsymbol{\mathsf{Q}_\mathsf{rep}}$ & $\boldsymbol{\mathsf{Q}_\mathsf{rej}}$ & $\boldsymbol{\mathsf{Q}_\mathsf{app}}$ & $\boldsymbol{\mathsf{Q}_\mathsf{rep}}$ & $\boldsymbol{\mathsf{Q}_\mathsf{rej}}$ & $\boldsymbol{\mathsf{Q}_\mathsf{app}}$ & $\boldsymbol{\mathsf{Q}_\mathsf{rep}}$ & $\boldsymbol{\mathsf{Q}_\mathsf{rej}}$ & $\boldsymbol{\mathsf{Q}_\mathsf{app}}$ & $\boldsymbol{\mathsf{Q}_\mathsf{rep}}$ & $\boldsymbol{\mathsf{Q}_\mathsf{rej}}$\cr
\midrule  
\textbf{Initialization} & 100.0\%& 0.0\%& 0.0\%& 97.0\%& 2.0\%& 1.0\%& 98.0\%& 2.0\%& 0.0\%& 84.0\%& 16.0\%&0.0\%\\
\textbf{Selection} & 100.0\%& 0.0\%& 0.0\%& 99.3\%& 0.7\%& 0.0\%& 100.0\%& 0.0\%& 0.0\%& 95.0\%& 4.0\%& 1.0\%\\
\textbf{Crossover} (P)& 99.6\%& 0.0\%& 0.4\%& 97.6\%& 2.4\%& 0.0\%& 95.6\%& 4.4\%& 0.0\%& {78.0\%}& 13.6\%& 8.4\%\\
\textbf{Mutation} (P)& 100.0\%& 0.0\%& 0.0\%& 100.0\%& 0.0\%& 0.0\%& 98.0\%& 2.0\%& 0.0\%& {92.3\%}& 3.3\%& 4.4\%\\
\textbf{Crossover} (S)& 100.0\%& 0.0\%& 0.0\%& 100.0\%& 0.0\%& 0.0\%& 99.6\%& 0.4\%& 0.0\%& {100.0\%}& 0.0\%& 0.0\%\\
\textbf{Mutation} (S)& 100.0\%& 0.0\%& 0.0\%& 100.0\%& 0.0\%& 0.0\%& 96.0\%& 3.7\%& 0.3\%& {99.9\%}& 0.0\%& 0.1\%\\
\bottomrule  
\end{tabular}  }
\label{netscience_format}
\end{table*}

\begin{table*}[!ht]
\centering
\caption{The validations for the output generated in different phases. The tested network is Erods. (P) refers to population-level reproduction and (S) refers to individual-level reproduction. Initialization refers to the sampling procedure.}
\resizebox{\textwidth}{!}{
\begin{tabular}{ccccccccccccc}  
\toprule  
\multirow{2}{*}{\textbf{Erods}}&  
\multicolumn{3}{c}{\textbf{Format Error} (GPT-4.0)}&\multicolumn{3}{c}{\textbf{Critical Error} (GPT-4.0)}&\multicolumn{3}{c}{\textbf{Format Error} (GPT-3.5)}&\multicolumn{3}{c}{\textbf{Critical Error} (GPT-3.5)}\cr  
\cmidrule(lr){2-4} \cmidrule(lr){5-7} \cmidrule(lr){8-10}\cmidrule(lr){11-13}  
& $\boldsymbol{\mathsf{Q}_\mathsf{app}}$ & $\boldsymbol{\mathsf{Q}_\mathsf{rep}}$ & $\boldsymbol{\mathsf{Q}_\mathsf{rej}}$ & $\boldsymbol{\mathsf{Q}_\mathsf{app}}$ & $\boldsymbol{\mathsf{Q}_\mathsf{rep}}$ & $\boldsymbol{\mathsf{Q}_\mathsf{rej}}$ & $\boldsymbol{\mathsf{Q}_\mathsf{app}}$ & $\boldsymbol{\mathsf{Q}_\mathsf{rep}}$ & $\boldsymbol{\mathsf{Q}_\mathsf{rej}}$ & $\boldsymbol{\mathsf{Q}_\mathsf{app}}$ & $\boldsymbol{\mathsf{Q}_\mathsf{rep}}$ & $\boldsymbol{\mathsf{Q}_\mathsf{rej}}$\cr
\midrule  
\textbf{Initialization} & 100.0\%& 0.0\%& 0.0\%& 98.0\%& 2.0\%& 0.0\%& 92.0\%& 8.0\%&0.0\%& 78.0\%& 14.0\%&8.0\%\\
\textbf{Selection} & 100.0\%& 0.0\%& 0.0\%& 92.2\%& 5.5\%& 2.3\%& 100.0\%& 0.0\%& 0.0\%& 91.1\%& 6.9\%& 2.0\%\\
\textbf{Crossover} (P)& 99.3\%& 6.3\%& 0.1\%& 98.6\%& 1.4\%& 0.0\%& 93.3\%& 6.3\%& 0.4\%& 74.6\%& 13.4\%& 12.0\%\\
\textbf{Crossover} (S)& 100.0\%& 0.0\%& 0.0\%& 100.0\%& 0.0\%& 0.0\%& 99.9\%& 0.1\%& 0.0\%& 100.0\%& 0.0\%& 0.0\%\\
\textbf{Mutation} (P)& 100.0\%& 0.0\%& 0.0\%& 99.3\%& 0.7\%& 0.0\%& 98.0\%& 1.6\%& 0.4\%& 93.0\%& 3.4\%& 3.6\%\\
\textbf{Mutation} (S)& 100.0\%& 0.0\%& 0.0\%& 100.0\%& 0.0\%& 0.0\%& 98.0\%& 1.6\%&0.4\%& 97.0\%& 1.3\%& 1.7\%\\
\bottomrule  
\end{tabular}  }
\label{erods_format}
\end{table*}

\begin{table*}[!ht]
\centering
\caption{The validations for the output generated in different phases. The tested network is Email. (P) refers to population-level reproduction and (S) refers to individual-level reproduction. Initialization refers to the sampling procedure.}
\resizebox{\textwidth}{!}{
\begin{tabular}{ccccccccccccc}  
\toprule  
\multirow{2}{*}{\textbf{Email}}&  
\multicolumn{3}{c}{\textbf{Format Error} (GPT-4.0)}&\multicolumn{3}{c}{\textbf{Critical Error} (GPT-4.0)}&\multicolumn{3}{c}{\textbf{Format Error} (GPT-3.5)}&\multicolumn{3}{c}{\textbf{Critical Error} (GPT-3.5)}\cr  
\cmidrule(lr){2-4} \cmidrule(lr){5-7} \cmidrule(lr){8-10}\cmidrule(lr){11-13}  
& $\boldsymbol{\mathsf{Q}_\mathsf{app}}$ & $\boldsymbol{\mathsf{Q}_\mathsf{rep}}$ & $\boldsymbol{\mathsf{Q}_\mathsf{rej}}$ & $\boldsymbol{\mathsf{Q}_\mathsf{app}}$ & $\boldsymbol{\mathsf{Q}_\mathsf{rep}}$ & $\boldsymbol{\mathsf{Q}_\mathsf{rej}}$ & $\boldsymbol{\mathsf{Q}_\mathsf{app}}$ & $\boldsymbol{\mathsf{Q}_\mathsf{rep}}$ & $\boldsymbol{\mathsf{Q}_\mathsf{rej}}$ & $\boldsymbol{\mathsf{Q}_\mathsf{app}}$ & $\boldsymbol{\mathsf{Q}_\mathsf{rep}}$ & $\boldsymbol{\mathsf{Q}_\mathsf{rej}}$\cr
\midrule   
\textbf{Initialization} & 100.0\%& 0.0\%& 0.0\%& 96.0\%& 4.0\%& 0.0\%& 98.0\%& 2.0\%& 0.0\%& 94.0\%& 6.0\%&0.0\%\\
\textbf{Selection} & 100.0\%& 0.0\%& 0.0\%& 92.2\%& 5.5\%& 2.3\%& 100.0\%& 0.0\%& 0.0\%& 91.1\%& 6.9\%& 2.0\%\\
\textbf{Crossover (P)}& 98.0\%& 2.0\%& 0.0\%& 94.3\%& 4.3\%& 1.4\%& 96.3\%& 3.7\%& 0.0\%& 71.6\%& 10.4\%& 18.0\%\\
\textbf{Crossover (S)}& 100.0\%& 0.0\%& 0.0\%& 100.0\%& 0.0\%& 0.0\%& 99.7\%& 0.2\%& 0.1\%& 100.0\%& 0.0\%& 0.0\%\\
\textbf{Mutation (P)}& 99.3\%& 0.6\%& 0.0\%& 99.7\%& 0.3\%& 0.0\%& 96.6\%& 3.4\%& 0.0\%& 89.8\%& 3.6\%& 6.6\%\\
\textbf{Mutation (S)}& 100.0\%& 0.0\%& 0.0\%& 100.0\%& 0.0\%& 0.0\%& 94.4\%& 4.9\%& 0.7\%& 100.0\%& 0.0\%& 0.0\%\\
\bottomrule  
\end{tabular}}
\label{email_format}
\end{table*}

The validation results across all networks in Tables \ref{netscience_format}-\ref{email_format} show that GPT-4.0 consistently outperforms GPT-3.5 in both format correctness and critical reasoning accuracy, particularly in population-level operations such as crossover and mutation. GPT-4.0 achieves near-perfect format compliance (${\mathsf{Q}_\mathsf{app}}$ and ${\mathsf{Q}_\mathsf{rep}}$ format error close to 100\%) and extremely low critical errors, even in more complex phases like crossover (P), where GPT-3.5 struggles with higher rejection rates and elevated critical errors. Fig. \ref{fig.netscience_moderate_error} also exhibits the similar result as format and critical check (obtained in Netscience). The ratio of $\mathsf{Q}_\mathsf{acc}$ in moderate error of GPT-4.0 is higher than that obtained by GPT-3.5. It can be concluded that the quality of LLMs output is promising but strongly dependent on model capability.

When comparing across networks (Netscience, Erdos, and Email), there is no significant difference in model performance trends. Both models behave consistently regardless of the underlying network structure, which is expected because the input provided to the LLMs is not the full graph, but only the solution representation (i.e., node indices). Therefore, the network topology does not directly influence the generation process.  Regarding the stages of evolutionary optimization, initialization generally produces fewer errors due to simpler output requirements (only sampling), while crossover and mutation (particularly with population-level way) are more error-prone. Finally, when comparing optimization strategies, individual-level reproduction (S) yields much lower errors than population-level (P), especially for GPT-3.5. This result indicates that generating individuals one at a time is significantly easier for less capable models, while GPT-4.0 handles both approaches well, showcasing its robustness in optimization tasks.

\begin{table*}[!ht]
\centering
\caption{Distributions of format, critical, and moderate errors in LLM-generated outputs across crossover and mutation operations with GPT-4.0  on three networks.}
\begin{tabular}{cccccccccccccc}  
\toprule  
\multirow{2}{*}{\textbf{Operation}} & \multirow{2}{*}{\textbf{Dataset}} &  
\multicolumn{3}{c}{\textbf{Format Error} (GPT-4.0)} & \multicolumn{3}{c}{\textbf{Critical Error} (GPT-4.0)} & \multicolumn{4}{c}{\textbf{Moderate Error} (GPT-4.0)}\cr  
\cmidrule(lr){3-5} \cmidrule(lr){6-8} \cmidrule(lr){9-12}
& & $\boldsymbol{\mathsf{Q}_\mathsf{app}}$ & $\boldsymbol{\mathsf{Q}_\mathsf{rep}}$ & $\boldsymbol{\mathsf{Q}_\mathsf{rej}}$ & $\boldsymbol{\mathsf{Q}_\mathsf{app}}$ & $\boldsymbol{\mathsf{Q}_\mathsf{rep}}$ & $\boldsymbol{\mathsf{Q}_\mathsf{rej}}$ & $\boldsymbol{\mathsf{Q}_\mathsf{app}}$ & $\boldsymbol{\mathsf{Q}_\mathsf{rep}}$ & $\boldsymbol{\mathsf{Q}_\mathsf{acc}}$ & $\boldsymbol{\mathsf{Q}_\mathsf{rej}}$\cr
\midrule  
\multirow{3}{*}{\textbf{Crossover}} 
& \textbf{Erods} & 99.3\%& 0.3\%& 0.4\%& 98.6\%& 1.4\%& 0.0\% & 50.0\% & 30.1\% & 19.3\% & 0.5\%\\
& \textbf{Email} & 100.0\% & 0.0\% & 0.0\% & 99.3\% & 0.7\% & 0.0\% &57.2\% & 24.0\% & 17.9\% & 0.8\%\\
& \textbf{Astro} & 99.3\% & 0.7\% & 0.0\% & 98.3\% & 1.7\% & 0.0\% & 55.8\% & 26.2\% & 18.1\% &0.8\%\\
\midrule  
\multirow{3}{*}{\textbf{Mutation}} 
& \textbf{Erods} & 100.0\%& 0.0\%& 0.0\%& 99.3\%& 0.7\%& 0.0\%& 80.8\% & 14.6\% & 4.5\% &0.1\%\\
& \textbf{Email} & 98.0\%& 2.0\%& 0.0\%& 94.3\%& 4.3\%& 1.4\% & 82.1\% & 14.3\% & 3.2\% &0.3\%\\
& \textbf{Astro} & 100.0\% & 0.0\% & 0.0\% & 100.0\% & 0.0\% & 0.0\% & 83.2\% & 14.0\% & 2.7\% &0.1\%\\
\bottomrule  
\end{tabular}  
\label{scala_analysis}
\end{table*}

\subsection{Scalability analysis}
The scalability of LLM-based operators is influenced by the input data, which varies across different stages.

\subsubsection{Initialization}
As illustrated in Table \ref{candidate_error}, the effectiveness of candidate selection is significantly influenced by the size of the dataset. \textbf{The input scale for LLMs during initialization for the candidate selection increases linearly with the dataset size}. For example, given a network of 10,000 nodes, we must input all their IDs with any specific metrics to LLMs for ranking, filtering, and sampling. It is not practical to input such large amounts of numerical data into LLMS and have them do such complex operations, not to mention the huge token overhead, meaning that LLMs may not be suitable for this data-intensive task. 


\subsubsection{Selection}
No matter what the dataset is, the input to LLMs is always $\{(S_1, f_1),...,(S_k, f_k)\}$. Thus, \textbf{the relevant factor to LLM-based selection is the population size.} A larger population size increases the amount of data LLMs need to process, but the common setting of this parameter is manageable to LLMs. Therefore, LLM-based selection has great applicability in this decision-making task given the promising result in Figure \ref{fig.selection_fitness}. 

\subsubsection{Crossover and Mutation}

{For these two phases, the input to LLMs is either $[\text{X}_1, \text{X}_2, ..., \text{X}_n]$ (individual-level) or $\{[\text{X}_{1}^{(1)},  \ldots, \text{X}_{n}^{(1)}], \ldots, [\text{X}_{1}^{(k)},  \ldots, \text{X}_{n}^{(k)}]\}$ (population-level). As such, \textbf{the input data amount, the main factor affecting LLMs performance, is dependent on the solution size and population size (only applicable to population-level optimization)}. The population size and solution size are empirical and will unnecessarily increase with the increase in the scale of datasets. Thus, we can conclude that the dataset scale and the LLM's performance are not directly related, which is demonstrated in Table \ref{scala_analysis} where the reliability of LLM-based EVO in the large dataset Astro is comparable to that in the two smaller networks. Evidence of LLMs sensitivity to hyperparameters can be found in Table \ref{Parameter_analysis}, where the population size is set to $\mathcal{P}_1=30$ and $\mathcal{P}_2=10$, and the solution size is set to $\mathcal{S}_1=10$ and $\mathcal{S}_2=5$. Reducing either the population size or the size of a single solution helps to minimize errors, a trend consistent for both crossover and mutation phases.} 

One potential factor affecting performance as dataset size increases is the representation of element IDs in the solution. For example, given two datasets with element counts of up to $10^3$ and $10^4$, the maximum number of digits required to represent element IDs is 4 and 5, respectively. This means that \textbf{the growth in element ID digits increases logarithmically rather than proportionally with dataset size.} Therefore, we conclude that the LLM-based EVO demonstrates excellent scalability in the reproduction phase with respect to structural manipulation. This is further supported by Table \ref{scala_analysis}, which shows no significant differences across the three types of checks between the Astro dataset and smaller datasets.

\begin{table*}[!ht]
\centering
\caption{Format, critical, and moderate error of LLM outputs during crossover and mutation on the Netscience dataset with varying population and solution size configurations.}
\begin{tabular}{ccccccccccccc}  
\toprule  
\multirow{2}{*}{\textbf{Netscience}}&  
\multicolumn{3}{c}{\textbf{Format Error} (GPT-4.0)}&\multicolumn{3}{c}{\textbf{Critical Error} (GPT-4.0)}&\multicolumn{4}{c}{\textbf{Moderate Error} (GPT-4.0)}\cr  
\cmidrule(lr){2-4} \cmidrule(lr){5-7} \cmidrule(lr){8-11} 
& $\boldsymbol{\mathsf{Q}_\mathsf{app}}$ & $\boldsymbol{\mathsf{Q}_\mathsf{rep}}$ & $\boldsymbol{\mathsf{Q}_\mathsf{rej}}$ & $\boldsymbol{\mathsf{Q}_\mathsf{app}}$ & $\boldsymbol{\mathsf{Q}_\mathsf{rep}}$ & $\boldsymbol{\mathsf{Q}_\mathsf{rej}}$ & $\boldsymbol{\mathsf{Q}_\mathsf{app}}$ & $\boldsymbol{\mathsf{Q}_\mathsf{rep}}$ & $\boldsymbol{\mathsf{Q}_\mathsf{acc}}$ & $\boldsymbol{\mathsf{Q}_\mathsf{rej}}$\cr
\midrule  
\textbf{C-}$\boldsymbol{(\mathcal{P}_1,\mathcal{S}_1)}$ & 96.7\%& 0.0\%& 3.3\%& 97.6\%& 2.4\%&0.0\%& 51.8\%& 27.0\%& 20.4\% &0.8\%\\
\textbf{C-}$\boldsymbol{(\mathcal{P}_1,\mathcal{S}_2)}$& 100.0\%& 0.0\%& 0.0\%& 99.4\%& 0.6\%& 0.0\%& 73.3\%& 18.9\%&7.6\%&0.2\%\\
\textbf{C-}$\boldsymbol{(\mathcal{P}_2,\mathcal{S}_1)}$ & 99.3\%& 0.7\%& 0.0\%& 99.3\%& 0.7\%&0.0\%& 93.2\%&4.8\%&1.5\%&0.5\%\\
\hline
\textbf{M-}$\boldsymbol{(\mathcal{P}_1,\mathcal{S}_1)}$ & 100.0\%& 0.0\%& 0.0\%& 100.0\%& 0.0\%& 0.0\%& 77.7\%& 18.4\%& 3.8\% &0.1\%\\
\textbf{M-}$\boldsymbol{(\mathcal{P}_1,\mathcal{S}_2)}$& 100.0\%& 0.0\%& 0.0\%& 99.6\%& 0.4\%& 0.0\%& 92.9\% & 6.3\%&0.8\%&0.0\%\\
\textbf{M-}$\boldsymbol{(\mathcal{P}_2,\mathcal{S}_1)}$ & 100.0\%& 0.0\%& 0.0\%& 98.0\%& 2.0\%& 0.0\%& 95.7\%& 3.8\%& 0.5\%&0.0\%\\
\bottomrule  
\end{tabular}  
\label{Parameter_analysis}
\end{table*}

\subsection{Validation of moderate error}

Moreover, we conducted an analysis to identify the specific moderate errors encountered by the LLM-based EVO during crossover and mutation. This investigation focuses on the population-level optimization, as the individual-level approach rarely produces such errors, as shown in Tables~\ref{netscience_format}-~\ref{email_format}. The full list of observed errors is provided in Tables \ref{error_message} and \ref{phase_check} of the main text. For clarity, we extract and re-index the relevant errors below:

\begin{itemize}
\item \textbf{Error 1}: The population after the operation remains the same as before.
\item \textbf{Error 2}: The size of the population fails to meet the requirement.
\item \textbf{Error 3}: The new solution does not preserve the original number of nodes in the original solution.
\item \textbf{Error 4}: The new solution contains duplicated elements.
\item \textbf{Error 5}: The new solution contains invalid nodes not found in candidate nodes (only applicable to mutation).
\end{itemize}

\begin{figure*}[!ht]
\centering
\includegraphics[width=0.7\textwidth]{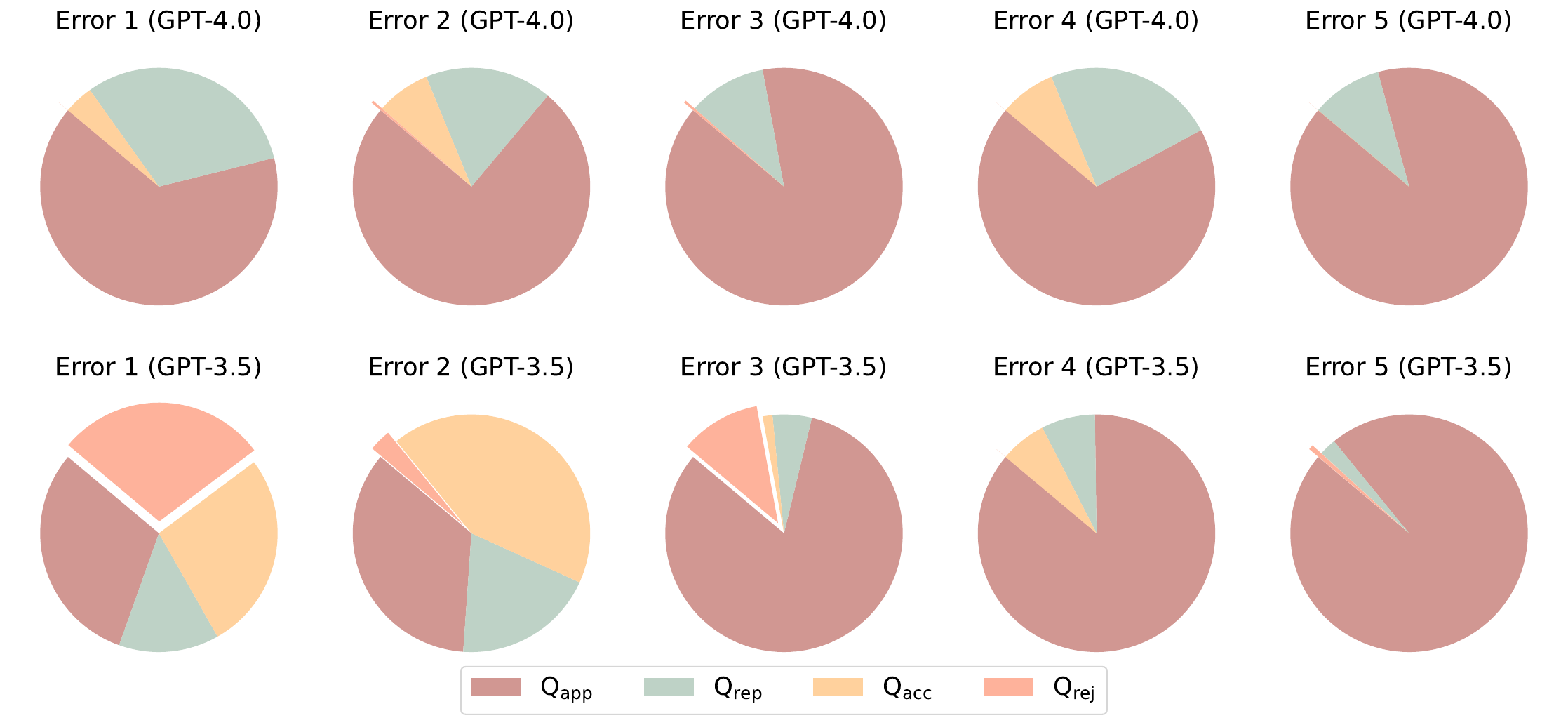}
\caption{{The observed moderate errors during population-level LLM-based mutation tested by GPT-4.0 and GPT-3.5. The tested network is Netscience $(|V|=379)$.}}
\label{moderate_error(mutation)}
\end{figure*}

These errors will be examined sequentially according to their index. When error 1 occurs, it will request a new operation from LLM. For the rest of errors, we will attach the corresponding message to the newly generated population and input them to LLMs for a repaired population. Figure \ref{moderate_error(mutation)} presents the results of moderate error in mutation. As observed, the selection of LLMs is one of the main factors of reliability. GPT-4.0's improvement over GPT-3.5 in managing mutation errors is evident across all error types except error 4, with a marked increase in the proportion of solutions that meet quality standards without the need for further modifications. On the other hand, GPT-3.5 is more likely to produce the unacceptable population with format or critical errors especially when trying to repair error 1 and error 3 while it rarely happens for GPT-4.0. It can also be found that the frequency of error varies, for example, error 3 and error 5 occur less frequently than others. Regardless of the LLM, error 1 happens most frequently although it can be repaired by GPT-4.0, it implies that LLMs sometimes will be overwhelmed with the data and does not work at all. The increased effectiveness in handling errors by GPT-4.0 suggests advancements in the model’s capability to generate more robust and reliable solutions, possibly due to improved understanding and processing of complex scenarios.

\begin{figure*}[!ht]
\centering
\includegraphics[width=0.7\textwidth]{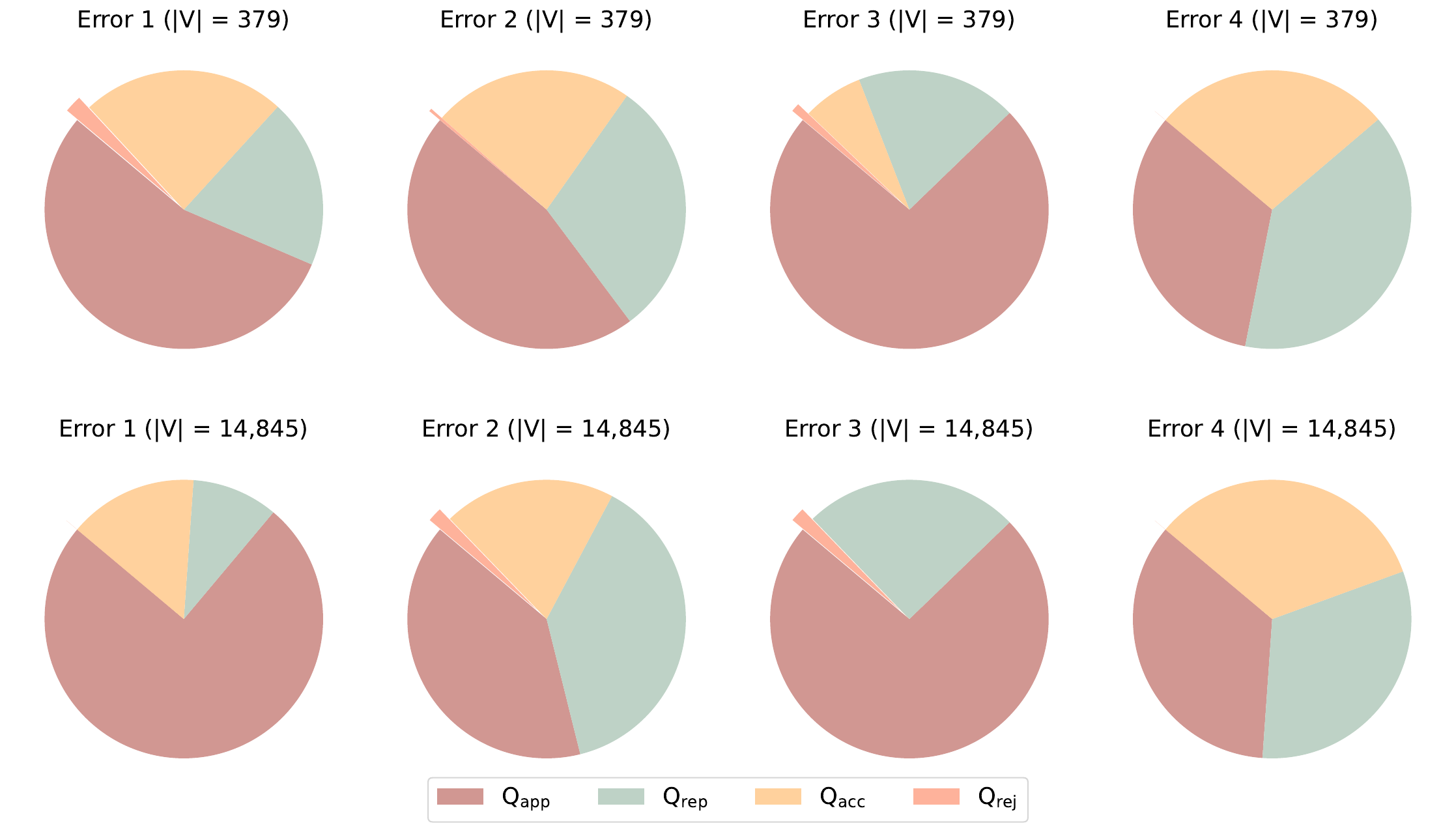}
\caption{The observed moderate errors during population-level LLM-based crossover. GPT-4.0 is used as the backbone LLM for testing. Two networks of different sizes are tested: Astro $(V=14,845)$ and Netscience $(|V|=379)$. }
\label{moderate_error(crossover)}
\end{figure*}

Figure \ref{moderate_error(crossover)} illustrates the moderate errors encountered during the crossover of two different-sized networks, Netscience and Astro. In contrast to mutation (Figure \ref{moderate_error(mutation)}), the $\mathsf{Q}_\mathsf{acc}$ ratio for LLM-based crossover is lower, suggesting that LLMs struggle more with crossover, possibly due to the need to manage two individuals compared to the single population in terms of operation.
We found that both networks exhibit a similar pattern despite their significant size difference, suggesting that \textbf{network size has little impact on the performance of LLM-based EVO regarding manipulating solutions} and highlighting its excellent scalability.

\subsection{Ablation study of repair mechanism}
\begin{figure*}[!ht]
\centering
\includegraphics[width=0.7\textwidth]{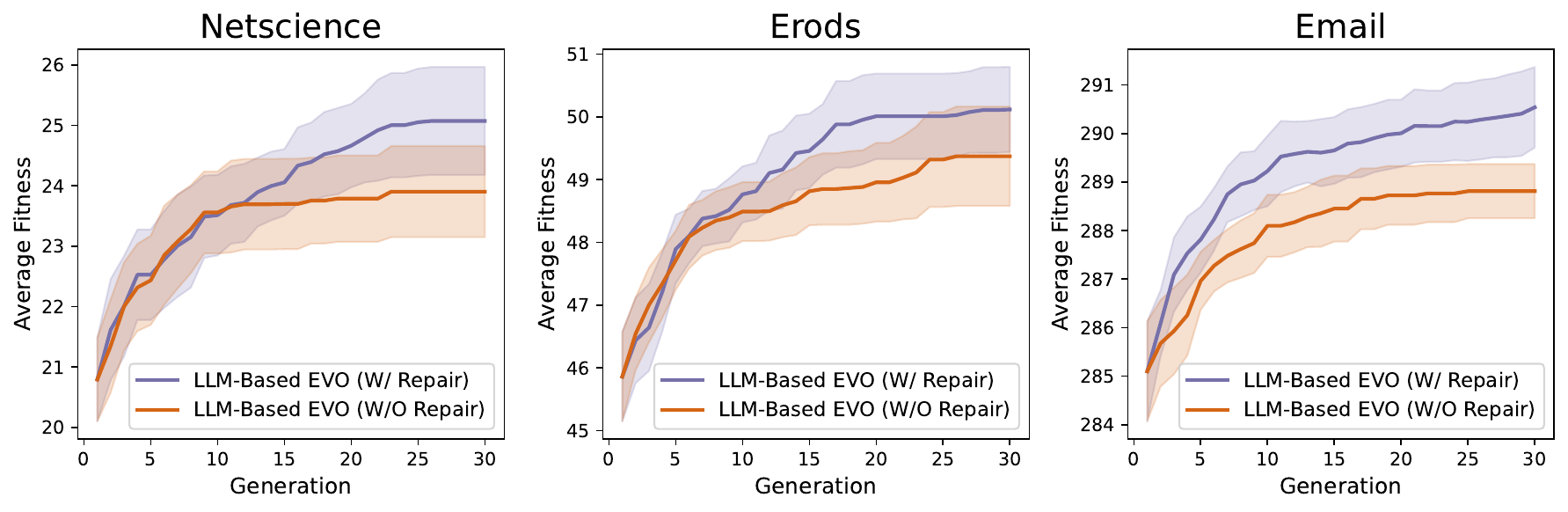}
\caption{{Comparison of average fitness across generations with and without the repair mechanism in LLM-based evolutionary optimization using GPT-4o on Netscience, Erods, and Email datasets.}}
\label{Ablation_study}
\end{figure*}

Figure \ref{Ablation_study} presents the ablation results of implementing a repair mechanism on the population-level LLM-based EVO across three datasets. We only test the moderate error since format error and critical error are so severe that the optimization cannot proceed as normal, and lead to consistently poor outcomes. For both datasets, \textbf{the inclusion of a repair mechanism consistently outperforms the absence of one}, as indicated by the higher average fitness achieved across generations. The observed stagnation in the scenarios without a repair mechanism indicates the impact of errors on optimization and suggests the importance of repair procedures.

\subsection{Computational overhead analysis}

\begin{figure*}[!ht]
\centering
\includegraphics[width=0.7\textwidth]{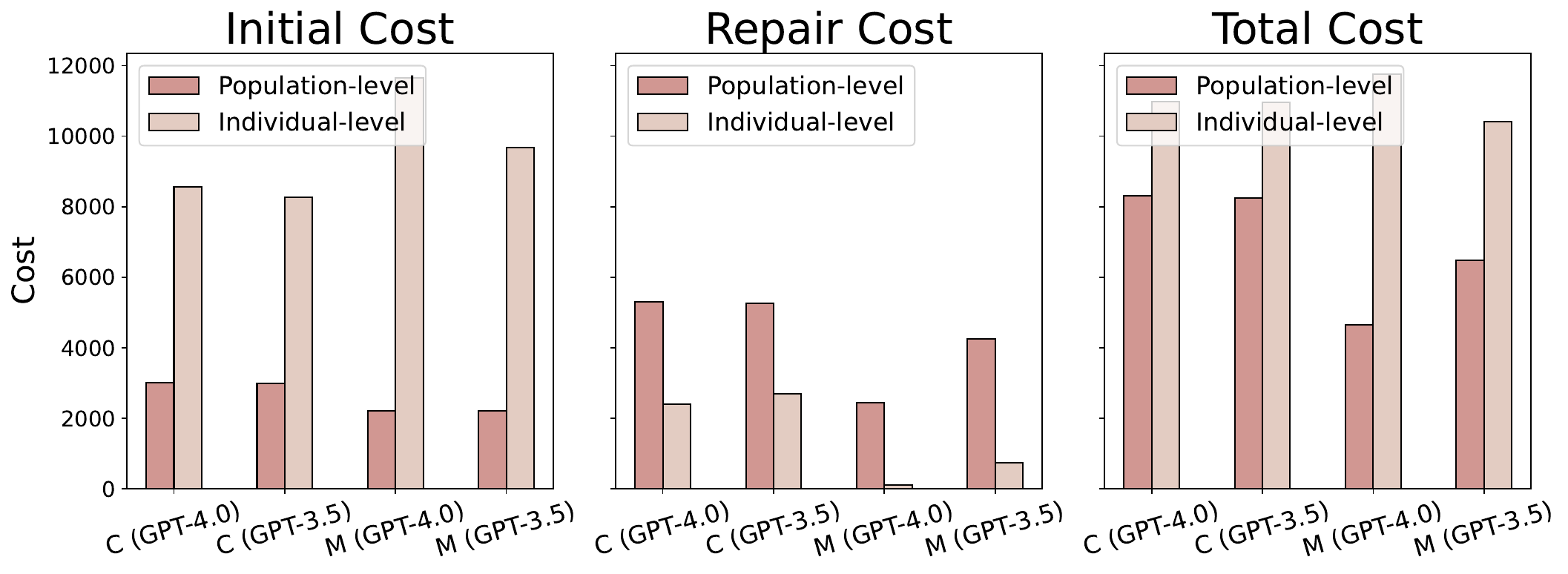}
\caption{Initial, repair, and total computational costs in terms of tokens for population-level and individual-level LLM-based evolutionary optimization using GPT-4.0 and GPT-3.5 during crossover (C) and mutation (M) operations.}
\label{fig.netscience_token_comparison}
\end{figure*}

Figure~\ref{fig.netscience_token_comparison} compares the token costs between individual- and population-level LLM-based EVO in Netscience. Initial cost refers to the tokens required for a single operation, while total cost includes the additional tokens for repairs. The results show that \textbf{population-level EVO incurs much lower costs than the individual-level EVO for both crossover and mutation in the entire process}. It can be seen that the repair costs for the population-level approach are higher than those for the individual-based method. Nevertheless, both theoretically and practically, the total costs of the population-level method are lower. As LLMs continue to advance, the accuracy of LLM-based EVO will improve, leading to reduced repair costs, which further shows the superiority of population-level optimization.

In the population-level optimization setting, we observe that GPT-4.0 incurs a higher initial token cost compared to GPT-3.5. This is primarily due to GPT-4.0’s greater reliability in generating the full number of individuals as specified in the prompt. While both models receive identical instructions, GPT-3.5 often produces incomplete populations, generating fewer individuals than requested, leading to shorter outputs and thus lower token usage. In contrast, GPT-4.0 tends to follow the prompt more precisely, resulting in more complete and token-heavy responses. This behavioral difference accounts for the higher initialization cost observed with GPT-4.0.

\section{Discussion}\label{Sec.dis}
In this section, we reflect on the limitations of LLM-based EVO and explore potential directions for extending LLM-based optimization to more complex and realistic scenarios.

\subsection{Toward Context-Aware Optimization via Visual Inputs}
\begin{figure}[!ht]
\centering
\includegraphics[width=0.49\textwidth]{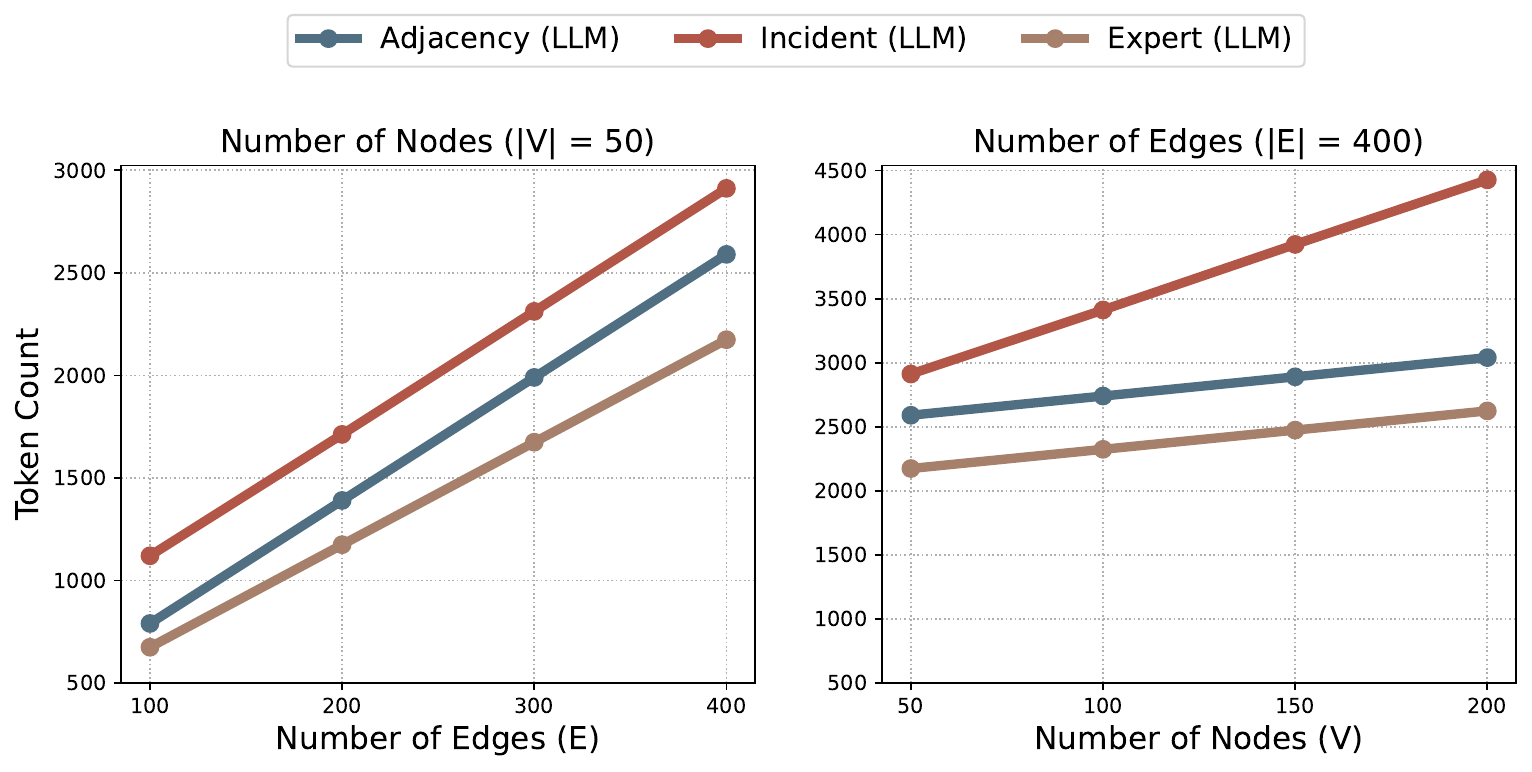}
\caption{Token cost comparison for different input formats when depicting networks of varying sizes in text.}
\label{fig.token}
\end{figure}

In this study, we concentrate on context-free optimization as a controlled setting to assess the reliability of LLMs as evolutionary optimizers. While this offers important insights, the broader objective is to enable LLMs to perform context-aware optimization. Achieving this requires the integration of graph representations into the input, allowing LLMs to leverage structural context and carry out more effective and informed optimization. However, integrating graph structures into LLMs inputs presents significant challenges, particularly for large-scale graphs: (1) \textbf{Computational Cost:} As illustrated in Figure \ref{fig.token}, the number of tokens required scales with network size across different input representations (e.g., adjacency, incidence, and expert ways) \cite{fatemi2023talk}. The token count increases linearly with the number of nodes and edges, significantly inflating the input size and computational burden. (2) \textbf{Efficacy:} Prior work \cite{fatemi2023talk, wang2024can} has shown that LLMs struggle to effectively interpret graph-structured data, often failing even on fundamental tasks. Moreover, their performance degrades rapidly as graph complexity increases. As such, current LLMs are not yet well-suited for directly handling complex real-world networks in the context-aware setting.

{Some problems that are computationally demanding for machines can be much more intuitive for humans, with combinatorial optimization being a prime example. When graph data is effectively visualized, humans can leverage their natural spatial and visual reasoning abilities to solve these problems more efficiently. With the emergence of multimodal large language models (MLLMs), we may be approaching a transformative moment in how such complex problems are addressed. Graphs represented as images, potentially with minimal high-order information loss thanks to advances in visualization, can now be interpreted by models capable of processing visual inputs, enabling machines to analyze graph structures in a more human-like way.}

{In addition, image-based inputs avoid the exponential growth in token count that characterizes text-based graph representations \cite{zhao2025visual}. This allows performance to remain efficient as the complexity of the network increases, with computational cost determined primarily by image size. These advantages suggest that MLLM-based evolutionary optimization could be a highly promising approach for solving combinatorial problems involving network structures, which we plan to investigate in future work.}

\subsection{Enhancing LLM-based EVO with Reasoning and Tools}
{While our study primarily utilizes GPT-3.5 and GPT-4.0 as representative LLMs, recent advancements in reasoning-augmented models open up promising avenues for enhancing the LLM-based EVO framework. These models are specifically designed to support deeper multi-step reasoning and could be well-suited for evolutionary optimization, where each generation involves iterative, structured decision-making. In our setting, operations such as selection, crossover, and mutation can be viewed as sequential reasoning tasks that build upon prior outputs. Reasoning-capable LLMs may offer improved stability and coherence across these steps, potentially leading to more consistent optimization trajectories.}

{In addition, tool-augmented LLMs like models equipped with access to external solvers, or plug-in functionalities offer another compelling direction for extending our approach. These tools could be leveraged to handle sub-tasks that are precision-critical or computationally intensive, such as constraint enforcement, population diversity maintenance, or fitness evaluation. Thus, it would allow the LLMs to focus on high-level strategic decisions (e.g., parent selection, mutation proposals), while offloading deterministic or repetitive computations to specialized modules. Such a hybrid framework could improve both the scalability and reliability of LLM-driven optimization.}

\section{Conclusion}\label{sec.conclusion}
In this work, we investigated the reliability and potential of large language models (LLMs) as evolutionary optimizers for network-structured combinatorial problems. To ensure robustness, we introduced a validation and repair mechanism to handle errors in model outputs and evaluated both individual-level and population-level optimization strategies. We also discussed future directions for enhancing LLM-based evolutionary optimization, including the integration of reasoning-capable models and external tools to handle sub-tasks requiring high precision. Additionally, we highlighted the limitations of context-free optimization and proposed the use of multimodal models capable of processing visual inputs as a promising solution for handling graph-structured data.

\bibliographystyle{IEEEtran}
\bibliography{zhao}

\vfill

\end{document}